%% file: main.tex
\documentclass[sigconf]{acmart}

\usepackage{booktabs}
\usepackage{caption}
\usepackage{amsthm}
\usepackage{algorithm}
\usepackage{algorithmic}
\usepackage{hyperref}
\usepackage{amsfonts}
\usepackage{subcaption}
\usepackage{booktabs} 
\usepackage{amsmath}
\usepackage{enumitem}
\usepackage{comment}
\usepackage{graphicx}
\usepackage{bm}
\usepackage{mathrsfs}
\usepackage{threeparttable}
\usepackage{multirow}
\usepackage{mathtools}
\usepackage{xurl} 
\usepackage{color}
\input{macros}

\setcopyright{none}

\usepackage{etoolbox}
\makeatletter
\patchcmd{\maketitle}{\@copyrightpermission}{
   \begin{minipage}{0.3\columnwidth}
     \href{https://creativecommons.org/licenses/by/4.0/}{\includegraphics[width=0.90\textwidth]{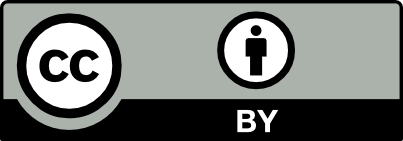}}
   \end{minipage}\hfill
   \begin{minipage}{0.7\columnwidth}
     \href{https://creativecommons.org/licenses/by/4.0/}{This work is licensed under a Creative Commons Attribution International 4.0 License.}
   \end{minipage}

   \vspace{5pt}
}{}{}

\makeatother

\newtheorem{assumption}{Assumption}

\AtBeginDocument{%
  \providecommand\BibTeX{{%
    \normalfont B\kern-0.5em{\scshape i\kern-0.25em b}\kern-0.8em\TeX}}}

\setcopyright{none}

\copyrightyear{2022}
\acmYear{2022}
\setcopyright{rightsretained}
\acmConference[KDD '22]{Proceedings of the 28th ACM SIGKDD Conference on Knowledge Discovery and Data Mining}{August 14--18, 2022}{Washington, DC, USA}
\acmBooktitle{Proceedings of the 28th ACM SIGKDD Conference on Knowledge Discovery and Data Mining (KDD '22), August 14--18, 2022, Washington, DC, USA}
\acmISBN{978-1-4503-9385-0/22/08}
\acmDOI{10.1145/3534678.3539299}

\begin{document}
\title{Learning Causal Effects on Hypergraphs}

\author{Jing Ma}
\affiliation{%
  \institution{University of Virginia}
  \city{Charlottesville, VA}
  \country{USA}
  }
\email{jm3mr@virginia.edu}

\author{Mengting Wan}
\affiliation{%
  \institution{Microsoft}
  \city{Redmond, WA}
  \country{USA}
  }
\email{mengting.wan@microsoft.com}

\author{Longqi Yang}
\affiliation{%
  \institution{Microsoft}
  \city{Redmond, WA}
  \country{USA}
  }
\email{longqi.yang@microsoft.com}

\author{Jundong Li}
\affiliation{%
  \institution{University of Virginia}
  \city{Charlottesville, VA}
  \country{USA}
  }
\email{jundong@virginia.edu}

\author{Brent Hecht}
\affiliation{%
  \institution{Microsoft}
  \city{Redmond, WA}
  \country{USA}
  }
\email{brent.hecht@microsoft.com}

\author{Jaime Teevan}
\affiliation{%
  \institution{Microsoft}
  \city{Redmond, WA}
  \country{USA}
  }
\email{teevan@microsoft.com}



\renewcommand{\shortauthors}{Jing Ma et al.}

\newcommand{\red}{\color{red}}
\newcommand{\mymodel}{\textbf{HyperSCI}}
\newcommand{\bigCI}{\mathrel{\text{\scalebox{1.07}{$\perp\mkern-10mu\perp$}}}}

\begin{abstract}
Hypergraphs provide an effective abstraction for modeling multi-way group interactions among nodes, where each hyperedge can connect any number of nodes. Different from most existing studies which leverage \textit{statistical dependencies}, we study hypergraphs from the perspective of \textit{causality}. Specifically, in this paper, we focus on the problem of individual treatment effect (ITE) estimation on hypergraphs, aiming to estimate how much an intervention {(e.g., wearing face covering)}
would causally affect an outcome {(e.g., COVID-19 infection)} of each individual node. Existing works on ITE estimation either assume that the outcome on one individual should not be influenced by the treatment assignments on other individuals (i.e., no \textit{interference}), or assume the interference only exists between pairs of connected individuals in an ordinary graph. We argue that these assumptions can be unrealistic on real-world hypergraphs, where higher-order interference can affect the ultimate ITE estimations due to the presence of group interactions. 
In this work, we investigate high-order interference modeling, and propose a new causality learning framework powered by hypergraph neural networks. Extensive experiments on real-world hypergraphs verify the superiority of our framework over  existing baselines. 
\vspace{-2mm}
\end{abstract}


%



\maketitle


\input{intro}

\input{problem}
\input{method}

\input{exp}

\input{related}

\vspace{-2mm}
\section{Conclusion}
\vspace{-1mm}
In this paper, we study an important research problem of individual treatment effect estimation with the existence of high-order interference on hypergraphs. We identify and analyze the influence of high-order interference in causal effect estimation. To address this problem, we propose a novel framework \mymodel, which estimates the ITEs based on representation learning. More specifically, \mymodel~ learns the representation of confounders, models the high-order interference with a hypergraph neural network module, then predicts the potential outcomes for each instance with the learned representations. We conduct extensive experiments to evaluate the proposed framework, where the results consistently validate the effectiveness of \mymodel~ in ITE estimation under different interference scenarios. 

\clearpage
\bibliographystyle{ACM-Reference-Format}
\bibliography{ref}

\clearpage
\input{appendix}
\end{document}

%% file: macros.tex
\newtheorem{theorem}{Theorem}[section]

\newtheorem{definition}[theorem]{Definition}

\newtheorem*{conjecture*}{Conjecture}
\newtheoremstyle{nonindented}{1ex}{1ex}{}{}{\bfseries}{.}{.5em}{}
\newtheoremstyle{indented}{1ex}{1ex}{\itshape\addtolength{\leftskip}{0.6cm}\addtolength{\rightskip}{0.6cm}}{}{\bfseries}{.}{.5em}{}
\theoremstyle{nonindented}
\theoremstyle{indented}
\theoremstyle{plain}

\renewcommand{\hat}{\widehat}
\renewcommand{\tilde}{\widetilde}

\newcommand{\eat}[1]{}

\newenvironment{lp*}{\begin{equation*}  \begin{array}{lll}}{\end{array}\end{equation*}}
\newcommand{\xhdr}[1]{\vspace{0.06in}\noindent{{\bf #1}}}
\setlist[itemize]{leftmargin=0.8\parindent,listparindent=\parindent,parsep=0.2\parskip,itemsep=0.02in,topsep=0.02in,after=\vspace{0in}}
\newcommand{\longeq}{\scalebox{3}[1]{=}}


%% file: intro.tex
\section{Introduction}\label{sec:intro}

\begin{figure}
\begin{subfigure}[b]{0.5\linewidth}
\centering
\includegraphics[width=\linewidth]{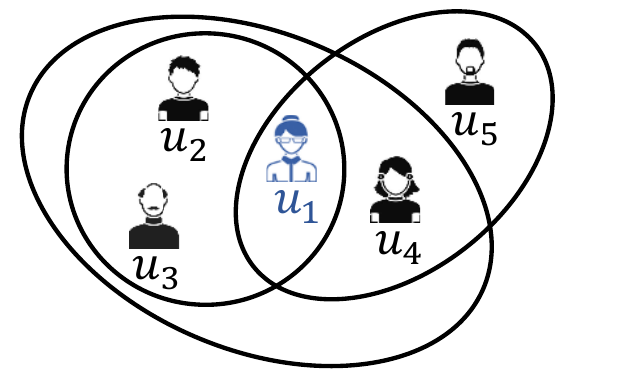}
\caption{Hypergraph}\label{fig:hypergraph}
\end{subfigure}
\qquad
\begin{subfigure}[b]{0.3\linewidth}
\centering
\includegraphics[width=\linewidth]{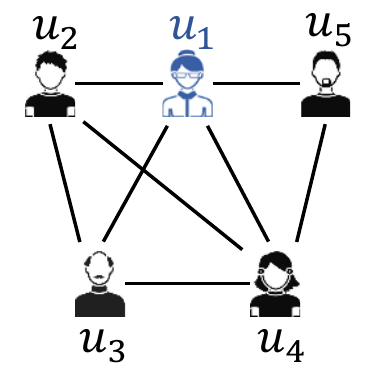}
\caption{Ordinary graph}\label{fig:ordinary-graph}
\end{subfigure}
\\
\begin{subfigure}[b]{\linewidth}
\centering
\includegraphics[width=\linewidth]{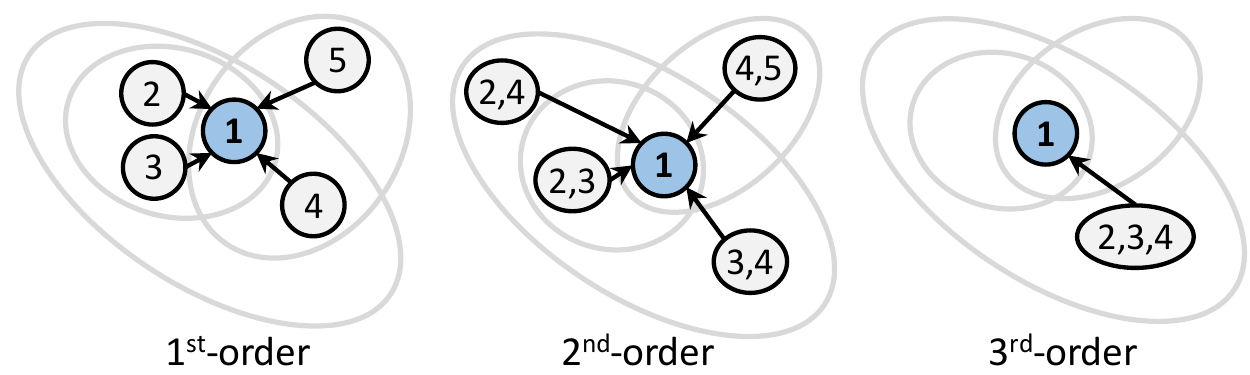}
\caption{First, second, and third-order interferences with $u_1$.}
\label{fig:high-order-interference}
\end{subfigure}
\vspace{-5mm}
\caption{(a) An illustrative example of group interactions on a hypergraph, where each circle represents a hyperedge (group); (b) An ordinary graph projected from this hypergraph; (c) Illustration of interferences with $u_1$ from its neighbors on the hypergraph. Note interference on (b) is pairwise (first-order only) while higher-order interference exists on the original hypergraph (a).
}
\vspace{-2mm}
\label{fig:interference}
\end{figure}

{Group interactions among individuals exist in a wide range of scenarios, 
e.g., massive gathering events, day-to-day group chats on WhatsApp or WeChat, and workplace interactions on Microsoft Teams or Slack channels. }
Although the conventional pairwise graph definition covers a vast number of applications (e.g., person-to-person physical contact networks or social networks \cite{brandes2013social}), it fails to capture the complete information of these group interactions (where each interaction may involve more than two individuals) \cite{bai2021hypergraph,feng2019hypergraph,yadati2018hypergcn}. The notion of the  \textit{hypergraph} can thus be introduced to address this limitation. 
Consider a hypergraph example that individuals are connected via in-person social events, each gathering event can be represented as a \textit{hyperedge} (Fig.~\ref{fig:hypergraph}). Each hyperedge can connect an arbitrary number of individuals, in contrast to an ordinary edge which connects exactly two nodes  (Fig.~\ref{fig:ordinary-graph}). 
While many studies have been devoted to utilizing such a generalized hypergraph structure to facilitate machine learning tasks \cite{bai2021hypergraph,feng2019hypergraph,yadati2018hypergcn,zhang2019hyper}, the majority were still executed at the statistical correlation level, e.g., 
{ predicting the COVID-19 infection risk on each individual (node) by capturing the \textit{correlations} between one's demographic information (node features), in-person group gathering history (hypergraph structure) and the infection outcomes (node labels).}
A critical limitation here is the lack of \textit{causality}, which is particularly important for understanding the impact of a policy intervention { (e.g., wearing face covering) on an outcome of interest (e.g., COVID-19 infection).} 
{ For individuals connected as in Fig.~\ref{fig:hypergraph}, one may ask ``how would each individual's face covering practice (treatment) \textit{causally} influence their infection risk (outcome)?''} 
Such a causal inference task requires constructing the counterfactual state of the same individual by holding all other possible factors constant except the treatment variable of interest. This is a particularly hard problem on hypergraph data, since the outcome of each individual is not only affected by their own confounding factors 
(e.g., one's health conditions and vaccine status) 
but also interfered by other individuals on the hypergraph
{ (e.g., face covering practice of other individuals who may physically contact the target individual through a gathering event).}

In this paper, we focus on learning causal effects on hypergraphs. We are specifically interested in estimating the individual treatment effect (ITE) under hypergraph interference from observational data. Our study is motivated by the following gaps:  
(i) \textit{Empirical constraints of randomized experiments.} One of the most reliable approaches for treatment effect estimation is randomized controlled trials (RCTs). Nevertheless, running RCTs is often expensive, impractical, even unethical \cite{goldstein2018ethical}, and they are especially difficult on graphs due to the dependencies among connected nodes \cite{ugander2013graph}.
(ii)  \textit{High-order interference on hypergraphs.} Our work focuses on the problem of ITE estimation, which aims to estimate the causal effect of a certain treatment (e.g., face covering practice) on an outcome (e.g., COVID-19 infection) for each individual. The classic ITE estimation is based on the Stable Unit Treatment Value (SUTVA) assumption \cite{fisher1936design,splawa1990application} that there is no interference \cite{tchetgen2012causal,hudgens2008toward} (i.e., spillover effect) among instances {(also referred to as \textit{units} in causal inference literature)}. That means the outcomes for any instance are not influenced by the treatment assignment of other instances. This assumption can be impractical in the real-world, thus resulting in flawed causal effect estimations, especially on graphs  where the interference among instances are ubiquitous \cite{ahluwalia2001moderating,yilmaz2002geographic,zeng2019exploration}. There have been many efforts addressing this problem \cite{aronow2017estimating,basse2018analyzing,imai2020causal,kohavi2013online,tchetgen2012causal,ugander2013graph,yuan2021causal,ma2021causal}, but most assume the interference only exists in a pairwise way on ordinary graphs (as shown in Fig.~\ref{fig:ordinary-graph}). This pairwise interference notion is insufficient to characterize the high-order interference that exists on hypergraphs. 
As shown in Fig.~\ref{fig:high-order-interference}, within a {gathering event} (hyperedge) between $u_1$, $u_2$ and $u_3$, an individual's ($u_1$) {infection outcome} can be affected by the \textit{first-order} interference from other individuals ($u_2\rightarrow u_1$ and $u_3\rightarrow u_1$) as well as the \textit{high-order} interference from the interactions among other individuals (the interaction between $u_2$ and $u_3$ may also act on influencing the exposure of the virus to $u_1$; consequently, $u_1$'s {infection risk} can be affected by this second-order interaction effect, i.e., $u_2\times u_3 \rightarrow u_1$). 
Notice that the number of such high-order interference items grows combinatorially as the size of a hyperedge increases, leading to a significant information gap between the original hypergraph and the projected pairwise ordinary graph (which accounts for the first-order interference only). This demands techniques capable of modeling high-order interference, but to the best of our knowledge, very little work has been done in this area.

In this paper, we propose a novel framework--- \textbf{C}ausal \textbf{I}nference under \textbf{S}pillover Effects in \textbf{Hyper}graphs (\textit{\mymodel})---to model high-order interference. At a high-level, 
this framework controls for the confounders and models high-order interference based on representation learning, then estimates the outcomes based on the learned representations. More specifically: (i) \emph{Controlling for Confounders.} Our framework is based on the widely accepted unconfoundedness assumption \cite{rubin1980randomization}, i.e., the confounders are contained in the observed features. With this assumption, we leverage representation learning techniques to capture and control for confounders from the features of each individual. Note as shown in previous works \cite{CFR}, the discrepancy between confounder distributions in the treatment group 
and the control group 
can lead to biases in causal effect estimations. Therefore, we also propose to use a representation balancing technique to mitigate the discrepancy between these two distributions.
(ii) \emph{Modeling High-order Interference.} Modeling high-order relationships can be challenging due to the complexity of enumerating multi-way interactions among nodes within each hyperedge. Historically, one may need to simplify the original hypergraph and approximate it through a series of projected ordinary graphs \cite{yoon2020much}. This obstacle is fortunately unblocked by the recent advances of hypergraph neural networks \cite{bai2021hypergraph,yadati2018hypergcn}. We extend this line of techniques to model interference by learning interference representations for each node. To learn the interference representations, the learned confounder representations and the treatment assignment are propagated via hypergraph convolution and attention operations. 
(iii) \emph{Outcome Prediction.} Based on the learned representations of confounders and interference, we predict the potential outcomes corresponding to different treatment assignments for each individual.
Overall, the main contributions of this work can be summarized as follows:

\begin{itemize}
    \item 
    We formalize the problem of ITE estimation under high-order interference on hypergraphs. To the best of our knowledge, it is the first work for this problem.
    \item 
    We propose a novel framework \mymodel~ for the studied problem. \mymodel~ models confounders and high-order interference via representation learning and hypergraph neural networks.
    \item 
    We validate the effectiveness of the proposed framework through extensive experiments and provide in-depth analysis on how it acts on different nodes and hyperedges.
\end{itemize}

%% file: problem.tex
\vspace{-2mm}
\section{Problem Definition and Analysis}
We provide the formal problem definition and a brief theoretical analysis of our studied problem in this section. A notation table is provided in Appendix A.
\begin{definition}
Suppose a set of individuals $\mathcal{V}=\{v_i\}_{i=1}^n$ are connected via hyperedges $\mathcal{E}=\{\mathbf{e}_k\}_{k=1}^m$, together these form a \textbf{hypergraph} $\mathcal{H}=\{\mathcal{V},\mathcal{E}\}$ with $n$ nodes and $m$ hyperedges, where each hyperedge can connect an arbitrary number of nodes. 
\end{definition}
The observational data on this hypergraph can be denoted as $\{\mathbf{X}, \mathcal{H}, \mathbf{T}, \mathbf{Y}\}$, where $\mathbf{X} =\{\mathbf{x}_i\}_{i=1}^n$, $\mathbf{T}=\{t_i\}_{i=1}^n$ and $\mathbf{Y}=\{y_i\}_{i=1}^n$ represent node features, treatment assignments, and observed outcomes, respectively. $\mathbf{H}=\{h_{i,e}\}\in\mathbb{R}^{n\times m}$ is an incidence matrix which describes the hypergraph structure of $\mathcal{H}$. $h_{i,e}=1$ if node $i$ is in hyperedge $e$, otherwise $h_{i,e}=0$. For ease of discussion, we consider the treatment assignment for each node as a binary variable in this study (i.e., $t_i\in\{0,1\}$), but our work can be extended to non-binary categorical variables and continuous variables.

{
\begin{definition}
The \textbf{potential outcome} \cite{rubin1980randomization} of the instance $i$ (denoted by $y_i^1$ or $y_i^0$) is defined as the realized value of outcome for instance $i$ under the treatment value $t_i=1$ or $t_i=0$. These potential outcomes can be instantiated via a transformation $Y^{T_i}_i=\Phi_Y(T_i, X_i, T_{-i}, X_{-i}, H)$.  $\Phi_Y$ can be regarded as a (non-deterministic) function to output potential outcomes, which takes each node's treatment assignment, node features, the information (treatment assignments and node features) of other nodes on the hypergraph, and the hypergraph structure as input\footnote{In this paper, we use non-bold, italicized, and capitalized letters (e.g., $X_i$) to denote random variables; non-bold lowercase letters (e.g., $t_i$) to denote observed values of a scalar; bold lowercase letters (e.g., $\mathbf{x}_i$) to denote observed values of a vector; bold capitalized letters (e.g., $\mathbf{X}$) to denote observed values of a matrix or a set.}, i.e., $y_i^{t_i}=\Phi_Y(t_i, \mathbf{x}_i, \mathbf{T}_{-i}, \mathbf{X}_{-i}, \mathbf{H})$, where the subscript ${-i}$ denotes all other nodes on $\mathcal{H}$ except $i$.
\end{definition}
}

Given the above preliminaries, 
we are ready to provide the formal definition of individual treatment effect on hypergraphs.

{
\begin{definition}
For each node $i$ on the hypergraph $\mathcal{H}$, the \textbf{individual treatment effect} (ITE) 
is defined by the difference between potential outcomes corresponding to  $t_i=1$ and $t_i=0$:
\begin{equation}
\begin{split}
    \tau(\mathbf{x}_i,\mathbf{T}_{-i},\mathbf{X}_{-i},\mathbf{H}) &=\mathbb{E}[Y_i^1-Y_i^0|X_i=\mathbf{x}_i,T_{-i}=\mathbf{T}_{-i},X_{-i}=\mathbf{X}_{-i},H=\mathbf{H}] \\
    &= \mathbb{E}[\Phi_Y(1,\mathbf{x}_i,\mathbf{T}_{-i},\mathbf{X}_{-i},\mathbf{H}) - \Phi_Y(0,\mathbf{x}_i,\mathbf{T}_{-i},\mathbf{X}_{-i},\mathbf{H})].
    \label{eq:ite}
\end{split}
\end{equation}
\end{definition}
}
{We clarify that the ITE in this paper is actually defined in the form of conditional average treatment effect (CATE), similar as \cite{ma2021causal,guo2020learning}. The expectation is taken over the potential outcome (output of $\Phi_Y$) of the instances with same node features $\mathbf{x}_{i}$ and “environmental information” (hypergraph structure $\mathbf{H}$, other nodes’ features $\mathbf{X}_{-i}$ and treatments $\mathbf{T}_{-i}$). 
The distribution of the output of $\Phi_Y$ is equivalent to the conditional distribution of the potential outcome conditioned on the parameters in $\Phi_Y$ with fixed values. For notation simplicity, we also denote $\tau_i=\tau(\mathbf{x}_i,\mathbf{T}_{-i},\mathbf{X}_{-i},\mathbf{H})$ in this paper.
}
Meanwhile, we introduce the notion of spillover effect in this work to assess the level of interference on hypergraphs.
\begin{definition}
The \textbf{spillover effect} of node $i$ under its treatment  $t_i$ and other nodes' treatment assignment $\mathbf{T}_{-i}$ on the hypergraph $\mathcal{H}$ is defined as:
\begin{equation}
    \delta_i = \mathbb{E}[\Phi_Y(t_i,\mathbf{x}_i,\mathbf{T}_{-i},\mathbf{X}_{-i},\mathbf{H}) - \Phi_Y(t_i,\mathbf{x}_i,\mathbf{0},\mathbf{X}_{-i},\mathbf{H})].
    \label{eq:spillover}
\end{equation}
\end{definition}

In this paper, given the observed data $\{\mathbf{X}, \mathcal{H}, \mathbf{T}, \mathbf{Y}\}$, we aim to estimate the ITE defined in Eq.~\ref{eq:ite} for each node in $\mathcal{H}$ with the existence of high-order interference defined in Eq.~\ref{eq:spillover}.

\subsection{Theoretical Analysis}
With the above definitions, we show that the ITE can be identifiable from the observational data under the following two assumptions.

Similar as the assumptions in other works of causal inference under network interference \cite{ma2021causal}, for each individual, we assume there exists a summary function capable of characterizing all the ``environmental'' information related to this node on the hypergraph.
Suppose there is a summary function $\mathtt{SMR}(\cdot)$: for each node $i$, $\mathtt{SMR}(\cdot)$ takes the hypergraph structure $\mathbf{H}$, the treatment assignment of other nodes $\mathbf{T}_{-i}$ and the features of these nodes $\mathbf{X}_{-i}$ as input, then maps them into a vector $\mathbf{o}_i$: 
\begin{equation}
      \mathbf{o}_i = \mathtt{SMR}(\mathbf{H}, \mathbf{T}_{-i}, \mathbf{X}_{-i}).
  \end{equation}
We use $H,X,T$ to denote the random variables for the hypergraph structure, features, and treatment assignment for any node. 
Then our first assumption can be formalized as below.
\begin{assumption}
(Expressiveness of summary function) 
For any node $i$, any values of $H, {X}_{-i}$, and ${T}_{-i}$, if the output of summary function $\mathbf{o}_i$ is determined, then the value of the potential outcomes $y_i^{1}$ and $y_i^{0}$ with feature $\mathbf{x}_i$ are also determined. 
\label{assumption:summary}
\end{assumption}

Our second assumption extends the unconfoundedness assumption \cite{rubin1980randomization} to the hypergraph interference setting. That is we assume conditioned on the above summary function, the observed features can capture all possible confounders.
\begin{assumption}
(Unconfoundedness) For any node $i$, given the node features, the potential outcomes are independent with the treatment assignment and summary of neighbors, i.e., $Y_i^1,Y_i^0 \bigCI T_i, O_i | {X}_i$. \label{assumption:unconfoundeness}
\end{assumption}

Based on the above assumptions, the identification of the expectation of potential outcomes $Y_i^{1}$ and $Y_i^{0}$ can be proved (here we take $Y_i^{1}$ as an example):
{ 
\begin{align}
&\mathbb{E}[Y^{1}_i|T_i=1, X_i=\mathbf{x}_i, T_{-i}=\mathbf{T}_{-i},{X}_{-i}=\mathbf{X}_{-i},H=\mathbf{H}] \\
\overset{(a)}{\longeq}&\mathbb{E}[\Phi_Y(T_i=1, X_i=\mathbf{x}_i, T_{-i}=\mathbf{T}_{-i},{X}_{-i}=\mathbf{X}_{-i},H=\mathbf{H})] \\\overset{(b)}{\longeq}& \mathbb{E}[\Phi_Y(T_i=1, X_i=\mathbf{x}_i, O_i=\mathbf{o}_i)] \\
\overset{(c)}{\longeq} &\mathbb{E}[\Phi_Y(T_i=1, X_i=\mathbf{x}_i, O_i=\mathbf{o}_i)|X_i=\mathbf{x}_i] \\
\overset{(d)}{\longeq} &\mathbb{E}[\Phi_Y(T_i=1, X_i=\mathbf{x}_i, O_i=\mathbf{o}_i)|X_i=\mathbf{x}_i, T_i=1,O_i=\mathbf{o}_i] \\
\overset{(e)}{\longeq} &\mathbb{E}[Y_i|X_i=\mathbf{x}_i, T_i=1,O_i=\mathbf{o}_i].
\end{align}
}

Here, the equation $(a)$ is based on the definition of potential outcome in this setting; $(b)$ is inferred from Assumption (1); $(c)$ is a straightforward derivation; $(d)$ is based on Assumption (2); and $(e)$ is based on the widely used consistency assumption \cite{rubin1980randomization}. Based on the above proof for the identification of potential outcomes, the identification of ITE can be straightforwardly derived.


%% file: method.tex
\section{The Proposed Framework}
\begin{figure*}[t]
\centering
     \includegraphics[width=.98\textwidth,height=1.3in]{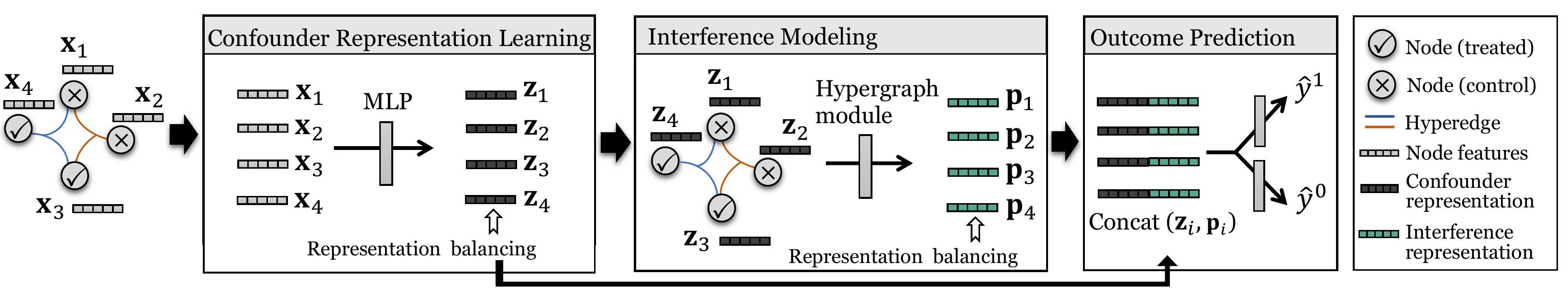}
     \vspace{-0.15in}
      \caption{An illustration of the proposed framework \mymodel, which includes three key components:  confounder representation learning, interference modeling, and outcome prediction. 
      }
      \vspace{-0.1in}
       \label{fig:model}
\end{figure*}

\begin{figure}[t]
\centering
     \includegraphics[width=.48\textwidth,height=1.3in]{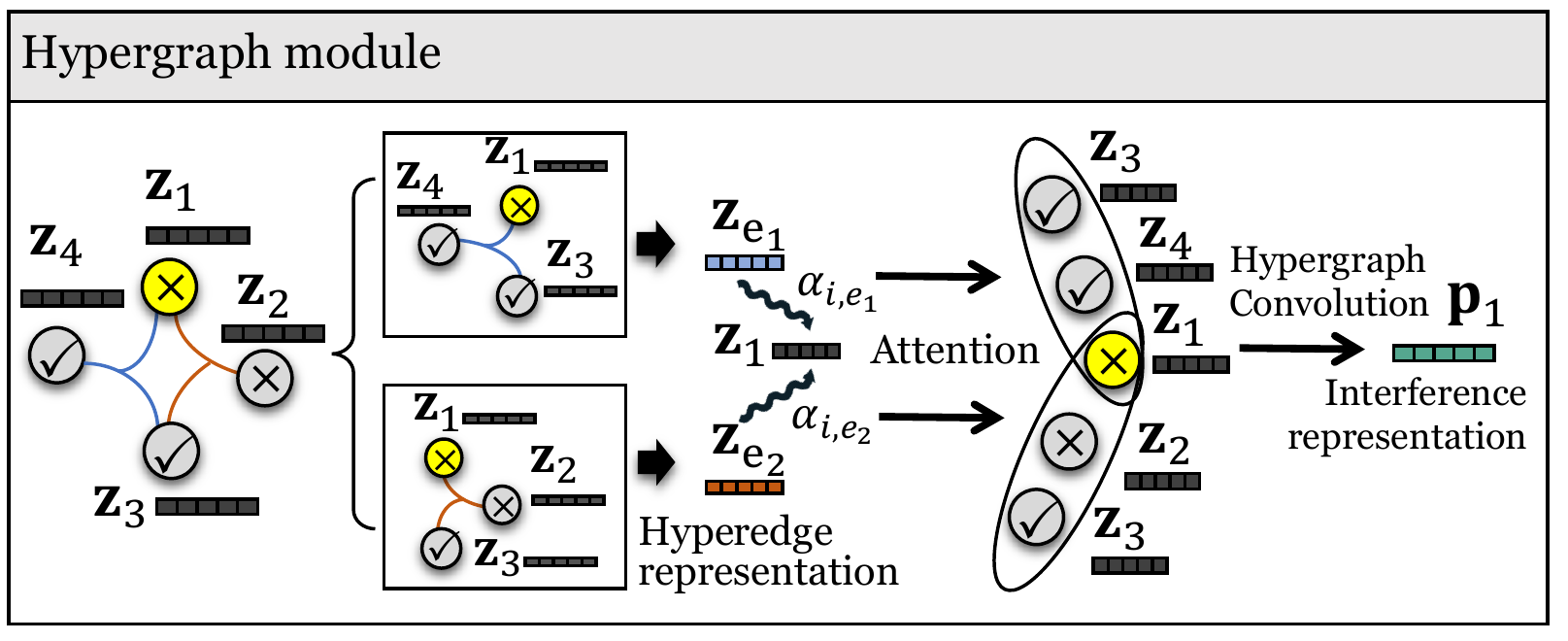}
     \vspace{-0.2in}
      \caption{A detailed illustration of the hypergraph module in the interference modeling component of \mymodel. Here we use the node $v_1$ (highlighted in yellow) as an example.
      }
     \vspace{-0.05in}
       \label{fig:model_hyper}
\end{figure}

Inspired by the previous theoretical analysis, we propose a novel framework \mymodel~to address the studied problem. This framework contains three components: confounder representation learning, interference modeling, and outcome prediction.
Holistically, we aim to learn an expressive transformation to summarize high-order interferences (Assumption~\ref{assumption:summary}), then take the interference representation, the confounder representation as well as the treatment assignment to estimate the expected potential outcome (Assumption~\ref{assumption:unconfoundeness}).
The illustration of \mymodel~ is shown in Fig.~\ref{fig:model}.

\subsection{Confounder Representation Learning}
We first encode the node features $\textbf{x}_i$ into a latent space via a multilayer perceptron (MLP) module, i.e., $\mathbf{z}_i=\text{MLP}(\mathbf{x}_i)$. This results in a set of representations $\mathbf{Z}=\{\mathbf{z}_i\}_{i=1}^n$, which is expected to capture all potential confounders, so the model can mitigate the confounding biases by controlling for the learned representation $\mathbf{z}_i$.

\paragraph{Representation Balancing.} Note a discrepancy may exist between the distributions of confounder representation $\mathbf{Z}$ in the treatment group and the control group, incurring biases in causal effect estimation, as shown in \cite{CFR,yao2018representation}. To minimize this discrepancy, we leverage the representation balancing technique by adding a discrepancy penalty to the loss function, where this discrepancy penalty can be calculated with any distribution distance metrics. In our implementation, we use the Wasserstein-1 distance \cite{CFR} between the representation distributions of treatment group and control group. 


\vspace{-2mm}
\subsection{Interference Modeling}
In this interference modeling component, we take the 
confounder representation ($\textbf{Z}$), the treatment assignment ($\mathbf{T}$), and the relational information on the hypergraph ($\textbf{H}=\{h_{i,e}\}$) as input, to capture the high-order interference for each individual. 
More specifically, we learn a transformation function $\Psi(\cdot)$ through a hypergraph module to generate the interference representations ($\mathbf{p}_i$) for each node $i$, i.e.,
$\mathbf{p}_i=\Psi(\mathbf{Z}, \mathbf{H}, \mathbf{T}_{-i}, t_i)$. As shown in Fig.~\ref{fig:model_hyper}, this module is implemented with a hypergraph convolutional network \cite{yadati2018hypergcn,bai2021hypergraph} and a hypergraph attention mechanism \cite{bai2021hypergraph,zhang2019hyper,ding2020more}, where the convolutional operator forms the skeleton of interference from hyperedges, and the attention operator enhances this mechanism by allowing flexible node contributions to each hyperedge. 

\vspace{-2mm}
\paragraph{Learning interference representations.}
To learn the representations which encode the interference in the hypergraph for each node, we propagate the treatment assignment and confounder representations with a hypergraph convoluntional layer. 
We first introduce a vanilla Laplacian matrix for the hypergraph $\mathcal{H}$:
\begin{equation}
\label{eq:laplacian}
    \mathbf{L} = \mathbf{D}^{-1/2}\mathbf{HB}^{-1}\mathbf{H}^\top\mathbf{D}^{-1/2}.
\end{equation}
Here $\mathbf{D}\in \mathbb{R}^{n\times n}$ is a diagonal matrix where each element represents the node degree (i.e., $\sum_{e=1}^m h_{i,e}$). $\mathbf{B}\in \mathbb{R}^{m\times m}$ is another diagonal matrix, where each element is the size of each hyperedge ($\sum_{i=1}^n h_{i,e}$).
Then we can define the hypergraph convolution operator as:
\begin{equation}
\label{eq: hgcn}
    \mathbf{P}^{(l+1)} = \text{LeakyReLU}\left(\mathbf{L}\mathbf{P}^{(l)}\mathbf{W}^{(l+1)}\right),
\end{equation}
where $\mathbf{P}^{(l)}$ represents the representations from the $l$-th layer in the hypergraph module. 
We feed the first layer with the previous confounder representation masked by the treatment assignment, i.e., $\mathbf{p}_i^{(0)}={t_i}*\mathbf{z}_i$,
where $*$ denotes element-wise multiplication.  
$\mathbf{W}^{(l+1)}\in \mathbb{R}^{d^{(l)}\times d^{(l+1)}}$ is the parameter matrix in the ($l$+1)-th layer, where $d^{(l)}$ and $d^{(l+1)}$ refer to the dimensionality of the interference representations in the $l$-th and ($l$+1)-th layers, respectively.

\paragraph{Modeling interference with different significance.}
Although the above convolution layer can pass interferences through hyperedges, it does not provide much flexibility to account for the significance of interference for different nodes through different hyperedges. 
In the aforementioned COVID-19 example, intuitively, those individuals who are active in certain group gathering events are more likely to influence or be influenced by others in these groups. To better capture this intrinsic relationship between nodes and hyperedges on a hypergraph, 
we leverage a hypergraph attention mechanism \cite{bai2021hypergraph,zhang2019hyper,ding2020more} to learn attention weights for each node and the corresponding hyperedges that contain this node. 

More specifically, we compute a representation for each hyperedge ($e$) by aggregating across its associated nodes ($\mathcal{N}_e$): $\mathbf{z}_{e}=\textsc{Agg}(\{\mathbf{z}_i~|~i\in \mathcal{N}_e\})$. Here, $\textsc{Agg}(\cdot)$ can be any aggregation functions (e.g., the mean aggregation). For each node $i$ and its associated hyperedge $e$, 
the attention score between a node $i$ and a hyperedge $e$ can be calculated as:
\begin{equation}
    \alpha_{i,e} = \frac{\text{exp}(\sigma(\text{sim}(\mathbf{z}_i \mathbf{W}_a,\mathbf{z}_e \mathbf{W}_a)))}{\sum_{k \in \mathcal{E}_i} \text{exp}(\sigma(\text{sim}(\mathbf{z}_i \mathbf{W}_a,\mathbf{z}_k \mathbf{W}_a)))},
\end{equation}
where $\sigma(\cdot)$ is a non-linear activation function, $\mathcal{E}_i$ denotes the set of hyperedges associated with the node $i$. Here we use $\mathbf{W}_a$ to denote a parameter matrix to compute the node-hyperedge attention. 
$\text{sim}(\cdot)$ is a similarity function, which can be implemented as:
\begin{equation}
    \text{sim}(\mathbf{x}_i, \mathbf{x}_j) = \mathbf{a}^\top[\mathbf{x}_i\|\mathbf{x}_j],
\end{equation}
where $\mathbf{a}$ is a weight vector, $[\cdot\|\cdot]$ is a concatenation operation. 

Then we use the attention scores to model the interference with different significance. Specifically, we replace the original incidence matrix $\mathbf{H}$ in Eq.~\ref{eq:laplacian} with an enhanced matrix $\tilde{\mathbf{H}}=\{\tilde{h}_{i,e}\}$, where $\tilde{h}_{i,e}=\alpha_{i,e}h_{i,e}$.
In this way, the interference from different nodes on the same hyperedge can be assigned with different importance weights, indicating different levels of contribution for interference modeling. We denote the final representations from the last convolution layer as $\mathbf{P}=\{\mathbf{p}_i\}_{i=1}^n$ and expect it to capture the high-order interference for each node.

\paragraph{Representation Balancing}
Similar to the coufounder representation learning module, we 
calculate a discrepancy penalty to reflect the difference between the distributions of interference representations in treatment and control groups. We sum up these two discrepancy penalties together to compute a representation balancing loss, denoted by $\mathcal{L}_b$.

\subsection{Outcome Prediction}
With the confounder representation $\mathbf{z}_i$ and the interference representation $\mathbf{p}_i$, we model the potential outcomes as:
\begin{align}
\hat{y}^1_i = f_1([\mathbf{z}_i \| \mathbf{p}_i]),\,\hat{y}^0_i = f_0([\mathbf{z}_i \| \mathbf{p}_i]),
\end{align}
where $f_1(\cdot)$ and $f_0(\cdot)$ are learnable functions to predict the potential outcome w.r.t. $t=1$ and $t=0$. We implement $f_1(\cdot)$ and $f_0(\cdot)$ with two MLP modules. Then the prediction for the observed outcome is obtained by $\hat{y}_i=\hat{y}^{t_i}_i$.
We optimize the model to minimize the following loss function: 
\begin{equation}
    \mathcal{L}=\sum_{i=1}^n (y_i-\hat{y}_i)^2 + \alpha\mathcal{L}_b + \lambda\|\mathbf{\Theta}\|^2,
\end{equation}
where the first term is the standard mean squared error, $\mathcal{L}_b$ is the representation balancing loss, $\mathbf{\Theta}$ represents the parameters in this neural network model. $\alpha$ and $\lambda$ are two hyperparameters which control the weights for the representation balancing loss and the parameter regularization term. The ITE for each instance $i$ can be estimated as: $\hat{\tau}_i=\hat{y}^1_i-\hat{y}^0_i$.

\subsection{Discussion} 
Here we revisit some implicit assumptions in the proposed framework. 
First, we assume that the interference for each node $i$ comes from its neighbors through the hypergraph structure. Here, the interference from neighbors that are multiple hops away can also be captured by stacking more hypergraph convoluntional layers.  
Second, for simplicity, we assume that the interference for each node only comes from other nodes with non-zero treatment assignment. 
Third, we assume that the representations of nodes in the same hyperedge are similar in the latent space. Besides, following \cite{bai2021hypergraph}, we assume that the representations of hyperedges are homogeneous with node representations. Nevertheless, we should still mention that this proposed framework is general and extendable, where the above assumptions can be further relaxed by enriching the hypergraph processing module.

%% file: exp.tex
\section{Experiments}
It is typically very hard to obtain the ground-truth counterfactual data as only one of the two potential outcomes can be obtained in the observational data. Hence, in this section, we follow a standard practice to evaluate the proposed framework and the alternative approaches on three semi-synthetic datasets. We aim to leverage as much real-world information as possible in the simulated environment. Our datasets are all based on real-world hypergraph data and we retain the treatment allocations as well as node features (covariates) if they are available. We simulate the outcome generation process to assess the true individual treatment effect (ITE), which eventually allows us to evaluate the performance of the ITE estimation from different causal inference approaches. 

\subsection{Dataset and Simulation}
We obtain the semi-synthetic data based on two publicly available hypergraph datasets (\textbf{Contact} \cite{mastrandrea2015contact,benson2018simplicial}, \textbf{Goodreads} \cite{wan2018item,wan2019fine}) and one large-scale proprietary web application dataset (\textbf{Microsoft Teams}).
We do not account for the temporal information of each hyperedge in our experiments and leave this as a future research direction instead.
In all three datasets, we discard extremely large hyperedges and keep those with no more than $50$ nodes only.\footnote{Note hyperedges with large size of nodes are usually less meaningful \cite{benson2018simplicial}.} 

\subsubsection{Outcome Simulation.} Given the treatment allocations $\mathbf{T}$, node features $\mathbf{X}$, and the hypergraph structure $\mathbf{H}$, the potential outcome of an individual $i$ can be simulated via

\begin{equation}
    y_i = f_{y,0}(\mathbf{x}_i) + \overbrace{\gamma f_t(t_i, \mathbf{x}_i)}^{\mathclap{\text{individual treatment effect (ITE)}}} + \underbrace{\beta f_s(\mathbf{T}, \mathbf{X}, \mathbf{H})}_{\mathclap{\text{hypergraph spillover effect}}} +~\epsilon_{y_i},
\end{equation}
where $f_{y,0}(\mathbf{x}_i)$ describes the outcome of instance $i$ when $t_i=0$ and without network interference, $f_t(\cdot)$ calculates the ITE of each instance,  $f_s(\cdot)$ calculates the spillover effect, and $\epsilon_{y_i}$ denotes the random noise from a Gaussian distribution $\mathcal{N}(0,1)$. 
We specify $f_{y,0}(\mathbf{x}_i)$ as a linear transformation of $\mathbf{x}_i$:
\begin{equation}
    f_{y,0}=\mathbf{w}_0\mathbf{x}_i,
\end{equation}
where $\mathbf{w}_0\sim \mathcal{N}(0,\mathbf{I}), \mathbf{w}_0\in \mathbb{R}^d$. Then we control the individual treatment effect ($f_t(t_i, \mathbf{x}_i)$) and the hypergraph spillover effect ($f_s(\mathbf{T}, \mathbf{X}, \mathbf{H})$) under two different settings:
\begin{enumerate}
    \item \textbf{Linear.}
      \begin{equation}
        f_t(t_i, \mathbf{x}_i) = \left\{
        \begin{aligned}
        & \mathbf{w}_1\mathbf{x}_i + \epsilon & \text{if } t_i=1\\
        & 0 & \text{if } t_i=0
        \end{aligned}\right.
    \end{equation}
Here $\mathbf{w}_1 \in \mathbb{R}^d$, and each element in $\mathbf{w}_1$ follows a Gaussian distribution. 
We generate $f_s$ as:
\begin{equation}
    f_s(\mathbf{T},\mathbf{X},\mathbf{H}) = \frac{1}{|\mathcal{E}_i|} \sum_{e\in\mathcal{E}_i} \sigma'(\frac{1}{|\mathcal{N}_e|}\sum_{j\in \mathcal{N}_e} t_j \times f_t(t_j, \mathbf{x}_j)).
\end{equation}
Here, $\sigma'(\cdot)$ is a function on the aggregation over each hyperedge. We implement it with an identity function by default.
    \item \textbf{Quadratic.}
     \begin{equation}
        f_t(t_i, \mathbf{x}_i) = \left\{
        \begin{aligned}
        & \mathbf{x}_i^\top\mathbf{W}_t\mathbf{x}_i + \epsilon & \text{if } t_i=1\\
        & 0 & \text{if } t_i=0
        \end{aligned}\right.
    \end{equation}
Here $\mathbf{W}_t \in \mathbb{R}^{d\times d}$, and each element in $\mathbf{W}_t$ follows a Gaussian distribution. 
We generate $f_s$ as:
\begin{equation}
    f_s(\mathbf{T},\mathbf{X},\mathbf{H}) = \frac{1}{|\mathcal{E}_i|} \sum_{e\in\mathcal{E}_i} \sigma'(\frac{1}{|\mathcal{N}_e|^2}(\mathbf{T}_e*\mathbf{X}_e)\mathbf{W}_t(\mathbf{T}_e*\mathbf{X}_e)^\top).
  \end{equation}    
  Here $\mathbf{X}_e$ and $\mathbf{T}_e$ are the feature matrix and treatment assignment of nodes contained in hyperedge $e$,  respectively. Here $*$ denotes element-wise multiplication.
\end{enumerate}

\subsubsection{Dataset Details.} We follow the above process to generate potential outcomes on all three datasets. Additional details about each dataset are provided as the follows.

\xhdr{Contact.} This dataset collects interactions recorded by wearable sensors among students at a high school \cite{mastrandrea2015contact,benson2018simplicial}, and includes 327 nodes and 7,818 hyperedges. 
Each node represents a person, and each hyperedge stands for a group of individuals are in close physical proximity to each other. This contact hypergraph data allows us to simulate a hypothetical question: ``how does one's face covering practice (treatment) causally affect their infection risk of an infectious disease (outcome)?''.
In each group contact, one may bring the virus to the surrounding environment, and thus affect other people's infection risk. Due to the lack of detailed information about each individual, apart from the potential outcome, we also generate the treatment ($t_i$) and the covariates ($\mathbf{x}_i$) as the follows:
\begin{equation}
    \mathbf{x}_i \sim \mathcal{N}(0,\mathbf{I}), {t}_{i} \sim Ber(\text{sigmoid}(\mathbf{x}_i\mathbf{v}_t)),
\end{equation}
where $\mathbf{I}$ is an $d\times d$ identity matrix, here we set $d=50$. $\mathbf{v}_t$ is a $d$-dimensional vector where each element inside follows a  Gaussian distribution. Eventually about $50\%\sim 60\%$ of the nodes are treated (${t}_{i}=1$) in our experiments.

\xhdr{GoodReads.} This dataset collects book information from the book review website GoodReads\footnote{https://www.goodreads.com/}, including the book title, authors, descriptions, reviews, and ratings \cite{wan2018item,wan2019fine}. We take each book in the \textit{Children} category as an instance. The bag-of-words of the book descriptions are used as the covariates of each book. 
Each hyperedge corresponds to each author and all books sharing the same author are in the same hyperedge.
The real-world book ratings are considered as treatment assignments: for each node $i$, we define $t_i=1$ if the rating score is larger than 3 and $t_i=0$ otherwise. We aim to study the causal effect of the rating score on the sales of each book. The ratings of each author's books can establish this author's overall reputation, and thus influence the sales of other books from the same author. The final processed dataset includes 57,031 nodes (where $40\%$ are treated) and 12,709 hyperedges. Note each book may have more than one author, and each author may have published multiple books.

\xhdr{Microsoft Teams.} We sampled 91,391 anonymized employees of a multinational technology company and collected their aggregated telemetry data on Microsoft Teams\footnote{\href{https://www.microsoft.com/en-us/microsoft-teams}{https://www.microsoft.com/en-us/microsoft-teams}}. Microsoft Teams is a workplace communication platform where users are allowed to create a group space (i.e., ``team'' or ``channel'') to enable public communication within each group. We are interested in how a user's usage of these group spaces causally affects their productivity. We process the treatment assignment into binary values by taking it as 1 if the employee has sent out at least one message in any of these group spaces during the first week of March, 2021; otherwise the treatment is assigned as 0. Each group space can be regarded as a hyperedge, where information can be shared via group discussions thus one's activeness on this platform may affect other individuals' outcomes in the same group. Employee demographics (e.g., office location, job description, work experience) were leveraged as the covariates. 

\begin{table}[]
\small
\centering
\setlength{\tabcolsep}{2.5pt}
 \caption{ITE estimation performance (mean $\pm$ standard error). ``CT", ``GR" and ``MS" stand for Contact, GoodReads and Microsoft Teams datasets, respectively.}
 \label{tab:baseline_linear}
\begin{tabular}{l|l|cc|cc}
\hline
              \multicolumn{1}{c|}{}                          &                           & \multicolumn{2}{c|}{Linear}                                                                     & \multicolumn{2}{c}{Quadratic}         \\\cline{3-6}
\multicolumn{1}{c|}{\multirow{-2}{*}{Data}} & \multirow{-2}{*}{Method} & $\sqrt{\epsilon_{PEHE}}$                                &  $\epsilon_{ATE}$                                                           &  $\sqrt{\epsilon_{PEHE}}$                 & $\epsilon_{ATE}$       \\
\hline
CT &  LR & $25.41$~\scriptsize{$\pm0.04$}         & $9.11$~\scriptsize{$\pm0.09$}   & $38.22$~\scriptsize{$\pm0.77$} & $20.28$~\scriptsize{$\pm0.38$}                     \\
 & CEVAE                     & $22.88$~\scriptsize{$\pm1.07$}         & $8.29$~\scriptsize{$\pm0.69$ } & $35.28$~\scriptsize{$\pm 0.75$} & $18.22$~\scriptsize{$\pm 0.76$}                  \\
 & CFR                       & $24.04$~\scriptsize{$\pm0.75$}         & $7.17$~\scriptsize{$\pm0.43$}  & $32.24$~\scriptsize{$\pm 1.01$} & $17.28$~\scriptsize{$\pm 0.75$}                \\
 & Netdeconf            & $10.22$~\scriptsize{$\pm0.47$}         & $4.29$~\scriptsize{$\pm0.13$} & $21.23$~\scriptsize{$\pm 0.72$} & $11.39$~\scriptsize{$\pm 0.74$ }      \\
 & GNN-HSIC               &   $7.42$~\scriptsize{$\pm0.39$}          & $2.06$~\scriptsize{$\pm 0.03$} & $16.28$~\scriptsize{$\pm 0.24$} & $7.28$~\scriptsize{$\pm 0.39$}   \\
 & GCN-HSIC               & $7.28$~\scriptsize{$\pm 0.44$}          & $2.08$~\scriptsize{$\pm 0.04$}  & $14.23$~\scriptsize{$\pm 0.20$} & $6.27$~\scriptsize{$\pm 0.15$} \\
 & \mymodel          & {\bm{$3.45$}~\scriptsize{\bm{$\pm 0.27$}}  }                  & {\bm{$1.39$}~\scriptsize{\bm{$\pm0.03$}}     }                                   & \bm{$9.20$}~\scriptsize{\bm{$\pm 0.09$}}  & \bm{$2.24$}~\scriptsize{\bm{$\pm 0.07$}}       \\
\hline
GR &  LR  & $23.01$~\scriptsize{$\pm 0.04$}         & $13.42$~\scriptsize{$\pm 0.12$}  & $48.56$~\scriptsize{$\pm 1.02$} & $31.19$~\scriptsize{$\pm 0.47$}                    \\
 & CEVAE                     & $22.69 $~\scriptsize{$\pm 0.03$}        & $12.49$~\scriptsize{$\pm 0.06$ }                 & $45.21$~\scriptsize{$\pm 3.10$} & $29.22$~\scriptsize{$\pm 0.44$}        \\
 & CFR                       & $20.30$~\scriptsize{$\pm 0.03$}         & $13.21$~\scriptsize{$\pm 0.09$ }                 & $41.72$~\scriptsize{$\pm 0.72$} & $26.28$~\scriptsize{$\pm 0.43$}        \\
 & Netdeconf                 & $18.39$~\scriptsize{$\pm 0.19$}         & $12.20$~\scriptsize{$\pm 0.03$}  & $35.18$~\scriptsize{$\pm 0.78$} & $21.20$~\scriptsize{$\pm 0.76$}            \\
 & GNN-HSIC                 & $17.20$~\scriptsize{$\pm 0.23$}         & $12.18$~\scriptsize{$\pm 0.13$}  & $27.22$~\scriptsize{$\pm 0.78$} & $16.87$~\scriptsize{$\pm 0.47$}             \\
 & GCN-HSIC                 & $16.01$~\scriptsize{$\pm0.20$}          & $12.06$~\scriptsize{$\pm 0.15$} & $25.42$~\scriptsize{$\pm 0.76$} & $16.28$~\scriptsize{$\pm 0.76$}     \\
 & \mymodel                      & \bm{$15.68$}~\scriptsize\bm{{$\pm0.21$}} & \bm{$11.81$}~\scriptsize\bm{{$\pm0.15$} }           & \bm{$19.23$}~\scriptsize\bm{{$\pm 0.44$}} & \bm{$13.33$}~\scriptsize{\bm{$\pm 0.27$}   }           \\
\hline
MS & LR    & $22.80$~\scriptsize{$\pm 0.64$}                   & $21.41 \pm$~\scriptsize{$0.74$}                                          & $414.17$~\scriptsize{$\pm 3.94$  }                    & $192.80$~\scriptsize{$\pm 2.97$}      \\
 &CEVAE           & $19.36$~\scriptsize{$\pm 0.80$}                    & $8.63$~\scriptsize{$\pm 0.78$}        & $315.01$~\scriptsize{$\pm2.53$} & $188.47$~\scriptsize{$\pm4.27$}            \\
 &CFR             & $25.23$~\scriptsize{$\pm  0.01$}                   & $18.28$~\scriptsize{$\pm  0.02$}          & $392.56$~\scriptsize{$\pm 4.33$} & $189.75$~\scriptsize{$\pm4.80$}  \\
 &Netdeconf       & $11.11$~\scriptsize{$\pm 0.01$}                  & $9.22$~\scriptsize{$\pm 0.03$ }      & $241.02$~\scriptsize{$\pm 2.32$ }                    & $147.29$~\scriptsize{$\pm1.04$}                                                 \\
 &GNN-HSIC    &$9.38$~\scriptsize{$\pm 0.44$}          & $6.91$~\scriptsize{$\pm  0.38$}             & $114.28$~\scriptsize{$\pm 3.62$}                     & $81.21$~\scriptsize{$\pm 2.53$}                               \\               
 &GCN-HSIC  &    $8.27$~\scriptsize{$\pm 0.41$} & $6.60$~\scriptsize{$\pm 0.48$} &   $109.57$~\scriptsize{$\pm3.85$}  & $77.75 $~\scriptsize{$\pm3.93$}  \\
 &\mymodel          & {\bm{$5.13$}~\scriptsize{\bm{$\pm 0.56$}}}               & {\bm{$4.46$}~\scriptsize{\bm{$\pm0.61$}} }    & \bm{$81.08$}~\scriptsize\bm{{$\pm 0.37$}  }            & \bm{$74.41$}~\scriptsize{\bm{$\pm0.42$}  }                                   \\
\hline
\end{tabular}
\end{table}

\subsection{Experiment Settings}
\subsubsection{Metrics}
We evaluate the performance of causal effect estimation through two standard metrics, including Rooted Precision in Estimation of Heterogeneous Effect ($\sqrt{\epsilon_{PEHE}}$) \cite{BART} and Mean Absolute Error ($\epsilon_{ATE}$) \cite{willmott2005advantages}. These metrics can be defined as follows:
\begin{equation}
    \sqrt{\epsilon_{PEHE}} = \sqrt{\frac{1}{n}\sum_{i\in[n]}(\tau_i-\hat{\tau}_i)^2}, \,\,\epsilon_{ATE} = |\frac{1}{n}\sum_{i\in [n]}{\tau_i}-\frac{1}{n}\sum_{i\in [n]}\hat{\tau}_i|.
\end{equation}
Lower $\sqrt{\epsilon_{PEHE}}$ or $\epsilon_{ATE}$ indicates better causal effect estimations.

\subsubsection{Baselines}
To investigate the effectiveness of our framework, we compare it with multiple state-of-the-art ITE estimation baselines. These baselines can be divided into the following categories:
\begin{itemize}
    \item \xhdr{No graph.} We compare the estimation results with traditional methods which do not consider graph data and spillover effects. These methods include outcome regression which is implemented by linear regression (\textbf{LR}), counterfactual regression (\textbf{CFR} \cite{CFR}), causal effect variational autoencoder (\textbf{CEVAE} \cite{CEVAE}). By comparing the proposed framework to these methods, we evaluate the effectiveness of modeling interference for ITE estimation.
    \item \xhdr{No spillover effect in ordinary graphs.} Although assuming no  spillover effect exists, the network deconfounder (\textbf{Netdeconf}) \cite{guo2020learning} captures latent confounders for ITE estimation by utilizing the network structure among instances. 
    \item \xhdr{Spillover effect in ordinary graphs.} We compare our framework with other ITE estimation baselines which can handle the pairwise spillover effect on ordinary graphs: 
    a node representation learning based method \cite{ma2021causal} estimates ITE under network interference, including two variants: (a) \textbf{GNN + HSIC}, which is based on graph neural network \cite{morris2019weisfeiler} and Hilbert Schmidt independence criterion (HSIC) \cite{gretton2005measuring}, and (b) \textbf{GCN + HSIC}, which is based on GCN \cite{kipf2016semi}. 
    
    To utilize the baselines which handle ordinary graphs, 
    we project the original hypergraph $\mathcal{H}$ to an ordinary graph $\mathcal{G}=\{\mathcal{V}, \mathcal{E}^p\}$ by setting $(v_i,v_j)\in \mathcal{E}^p$ if $v_i$ and $v_j$ are contained in at least one common hyperedge in $\mathcal{H}$. By comparing \mymodel~to the above baselines, we are able to evaluate the benefits of modeling high-order interferences on the original hypergraph.
\end{itemize}

\subsubsection{Setup} We randomly partition all datasets into 60\%-20\%-20\% training/validation/test splits. All the results are averaged over ten repeated executions. Unless otherwise specified, we set the hyperparameters as $\alpha=0.001$, $\beta=1.0$, $\gamma=1.0$, $\lambda=0.01$, the dimension for confounder representation and interference representation both as $64$. We use ReLU as the  activation function, and use an Adam optimizer. By default, the interference modeling component contains one hypergraph convolutional layer.

\subsection{ITE Estimation Performance}
We include the results for the ITE estimation task in Table~\ref{tab:baseline_linear}. From this table we 
observe that the proposed framework outperforms all the baselines under both linear and quadratic outcome simulation settings. We attribute these results to the fact that \mymodel~ utilizes the  relational information in hypergraph to model the high-order interference, and thus mitigates the influence of the spillover effect on ITE estimation performance. 
Compared with other baselines, the methods which incorporate the pairwise network interference (\textbf{GCN-HSIC} and \textbf{GNN-HSIC}), as well as \textbf{Netdeconf} which utilizes the network structure for ITE estimation, perform better than those baselines which do not take advantage of the relational information (\textbf{LR}, \textbf{CEVAE}, \textbf{CFR}). 

\begin{figure}[t]
\centering
    \begin{subfigure}[b]{0.235\textwidth}
        \centering
        \includegraphics[height=0.95in]{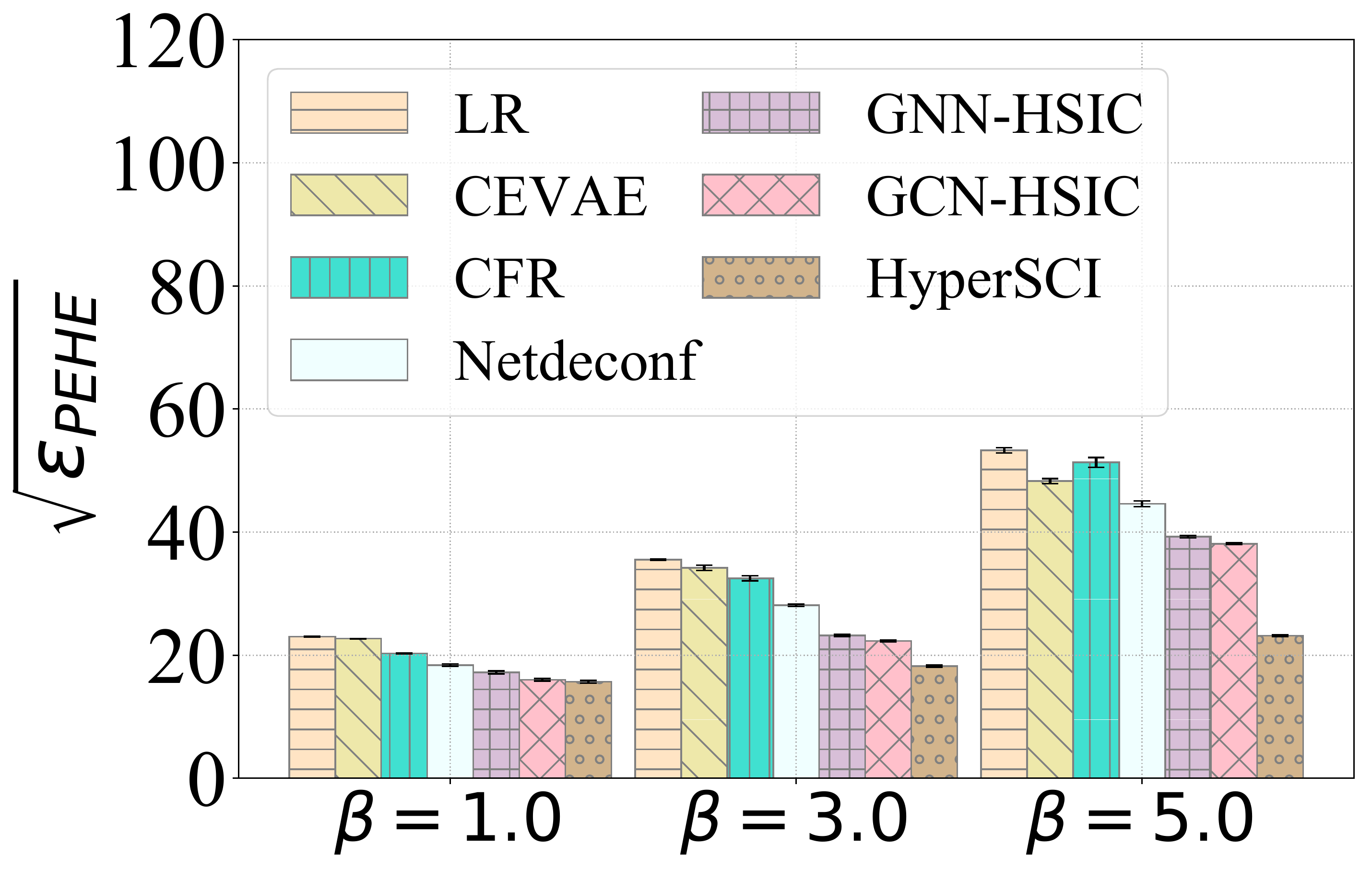}
        \caption{$\sqrt{\epsilon_{PEHE}}$}
    \end{subfigure}
  \begin{subfigure}[b]{0.235\textwidth}
        \centering
        \includegraphics[height=0.95in]{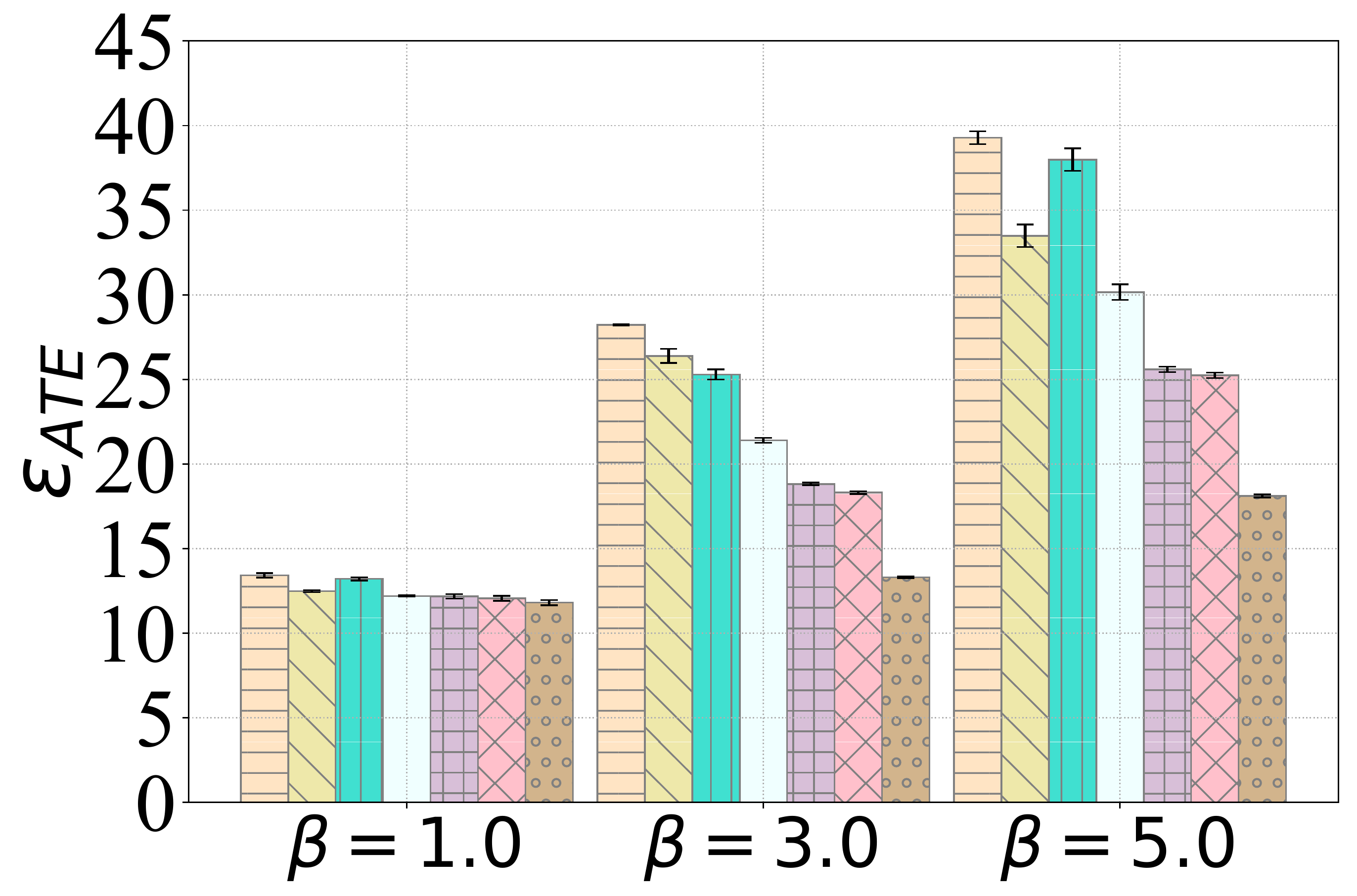}
        \caption{$\epsilon_{ATE}$}
    \end{subfigure}
   \vspace{-7mm}
  \caption{Comparison of the performance of ITE estimation under different values of $\beta$ in linear setting on GoodReads.}
  \label{fig:ITE}
\vspace{-1mm}
\end{figure}

We also vary the hyperparameter ($\beta$) which controls the significance of hypergraph spillover effect in the outcome simulation and report the ITE estimation results in Fig.~\ref{fig:ITE}.
As $\beta$ increases, i.e., the outcome is more heavily influenced by interference, larger performance gains can be observed from the proposed framework (\mymodel) against baselines. 
This observation further validates the effectiveness of our framework in modeling the interference for enhancing the performance of ITE estimation.

\subsection{Ablation Study}

\begin{figure}
\begin{subfigure}[b]{0.22\textwidth}
        \centering
        \includegraphics[height=.8in]{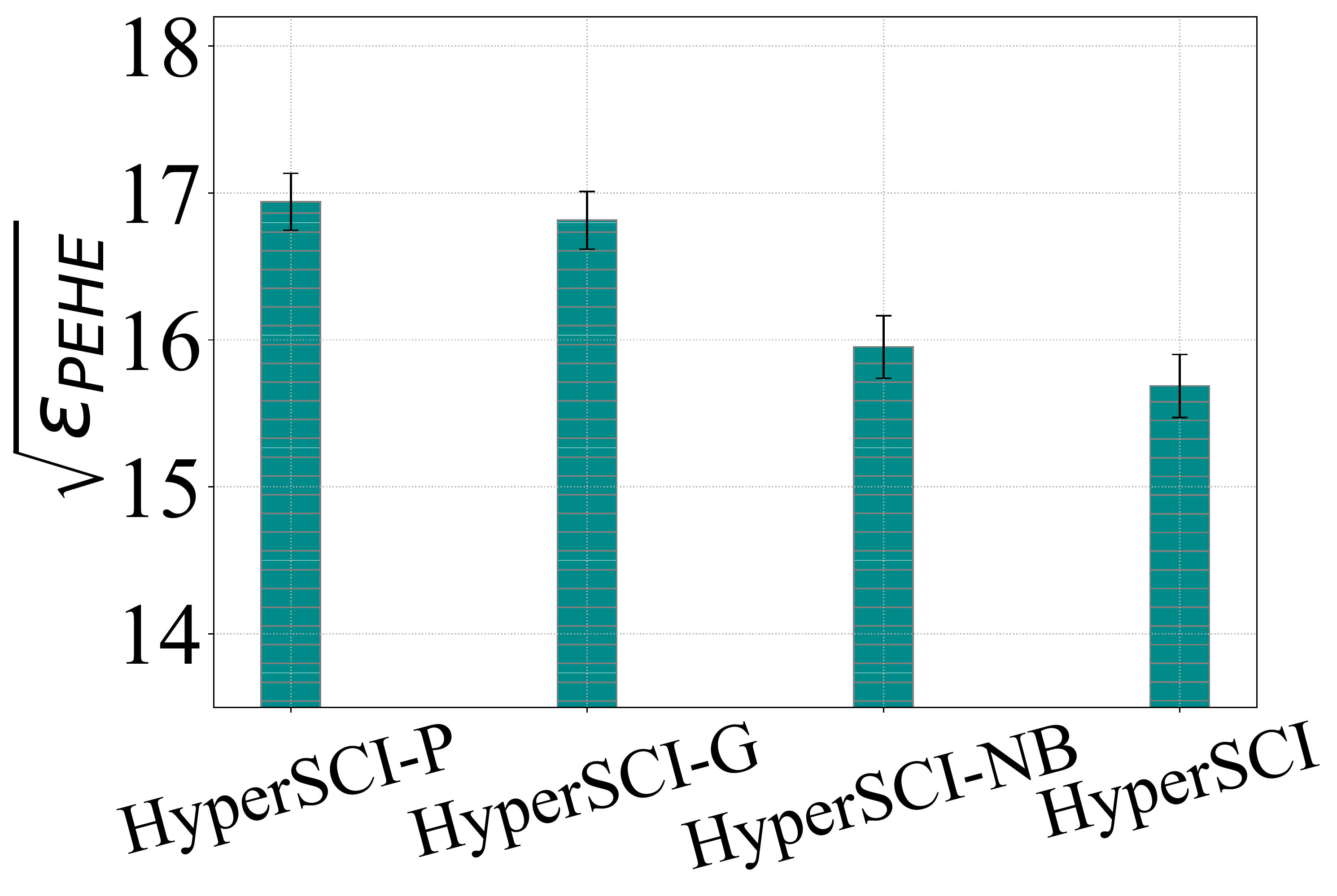}
        \caption{GoodReads, $\sqrt{\epsilon_{PEHE}}$}
    \end{subfigure}
    \begin{subfigure}[b]{0.22\textwidth}
        \centering
        \includegraphics[height=.8in]{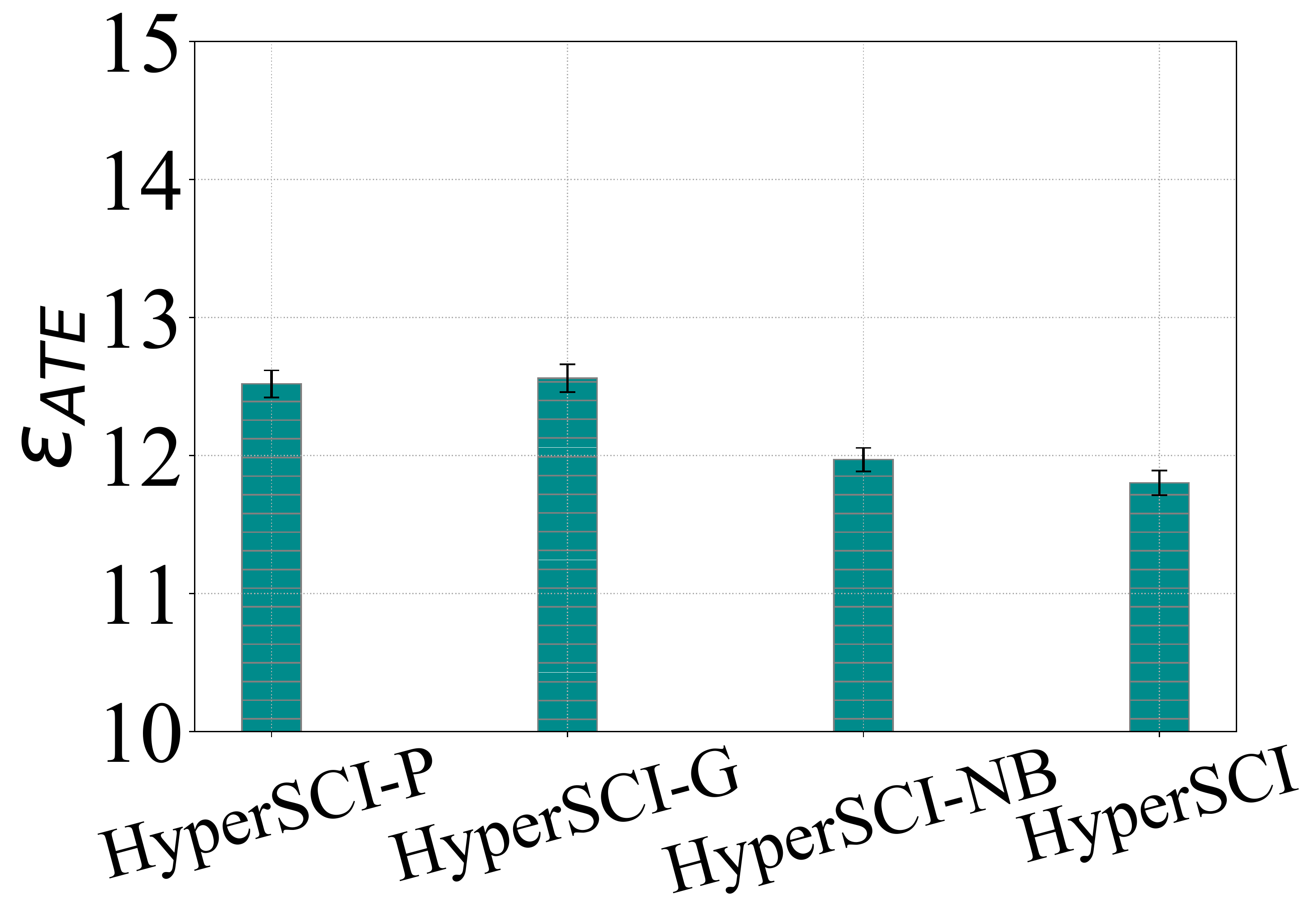}
        \caption{GoodReads, $\epsilon_{ATE}$}
    \end{subfigure}
    \begin{subfigure}[b]{0.22\textwidth}
        \centering
        \includegraphics[height=.8in]{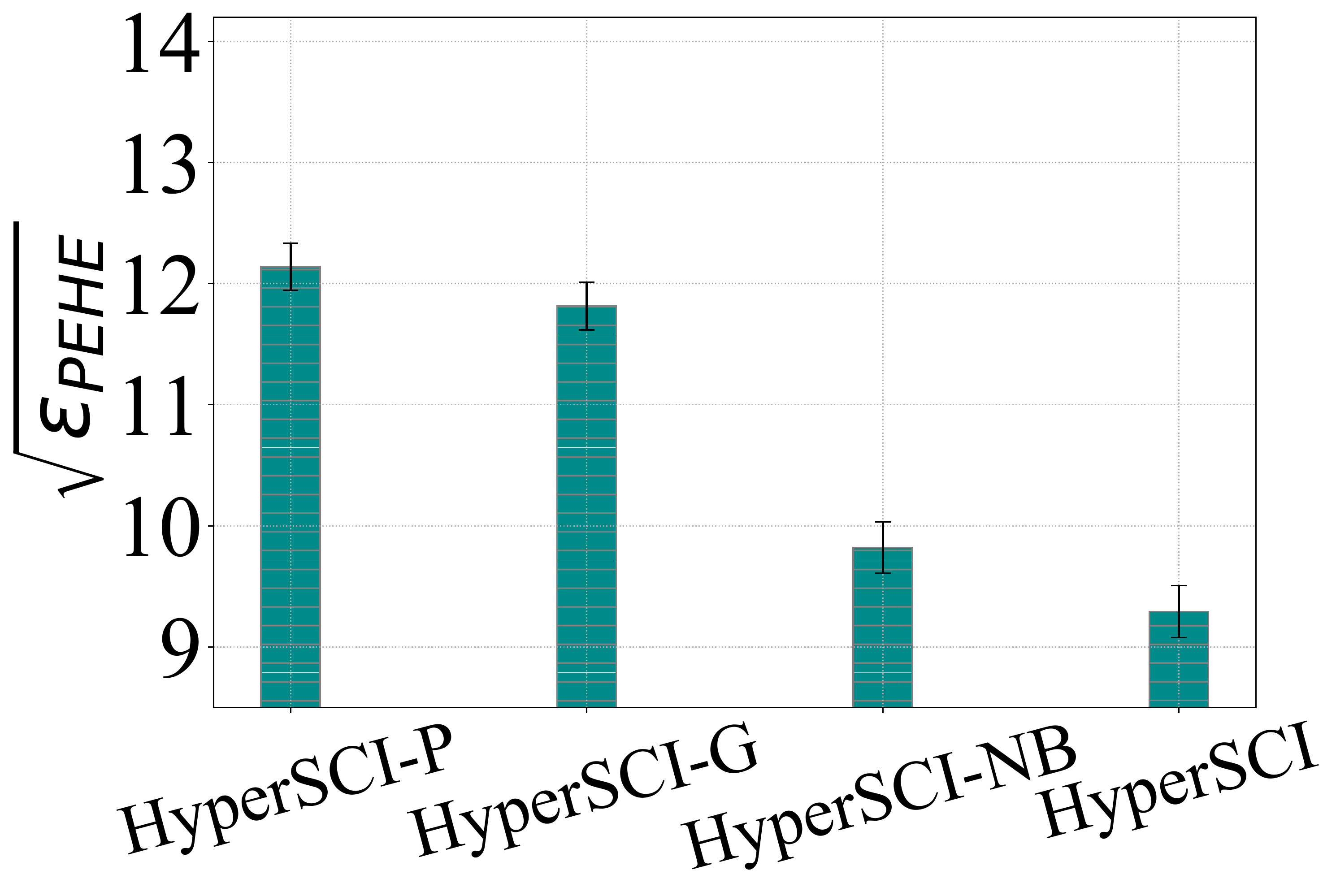}
        \caption{Contact, $\sqrt{\epsilon_{PEHE}}$}
    \end{subfigure}
  \begin{subfigure}[b]{0.22\textwidth}
        \centering
        \includegraphics[height=.8in]{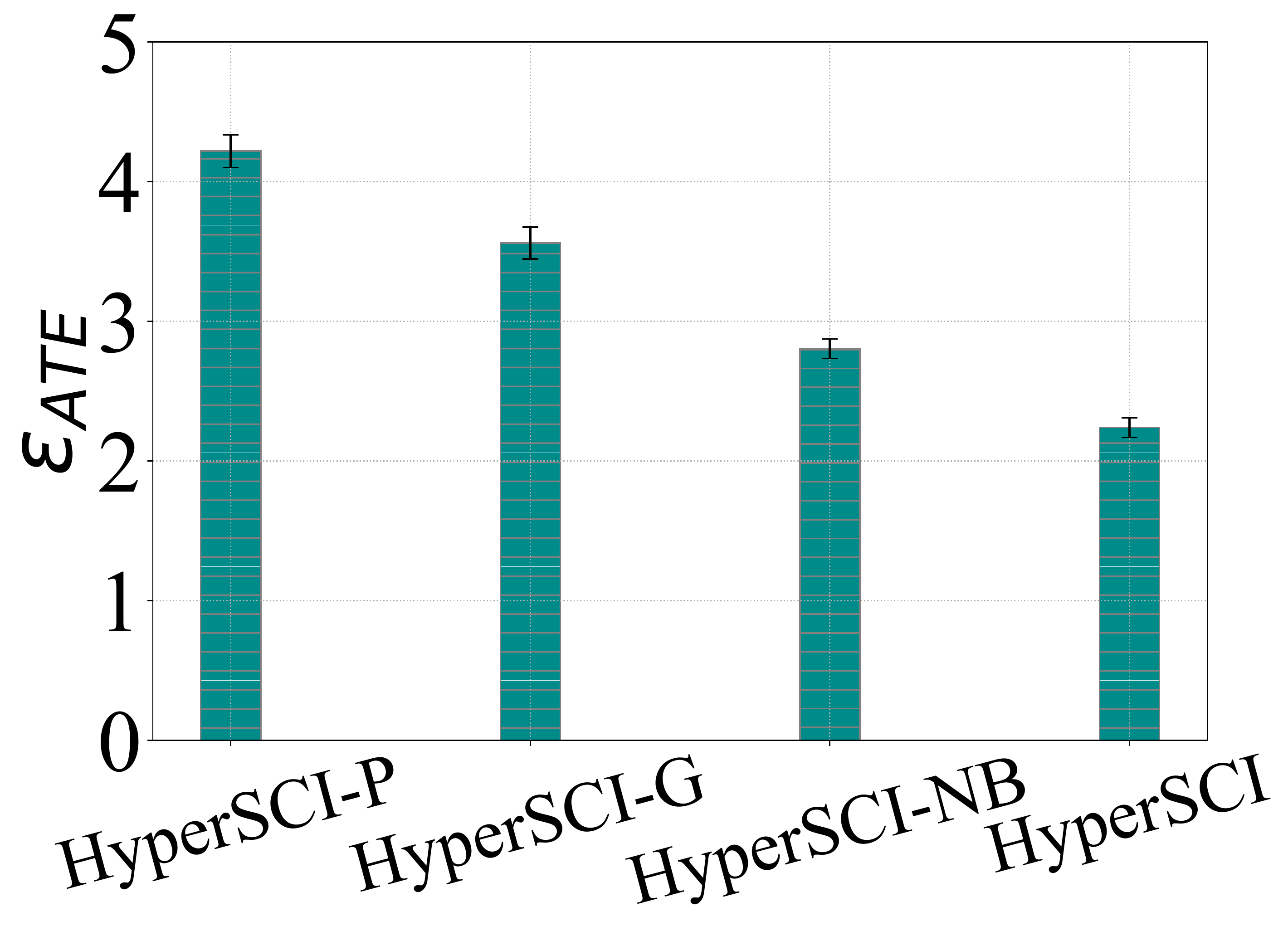}
        \caption{Contact, $\epsilon_{ATE}$}
    \end{subfigure}
    \vspace{-1mm}
  \caption{Ablation studies of different variants of our framework \mymodel. Results (mean and standard error) are reported under the linear setting but similar patterns can be found under the quadratic setting and on all datasets. }
  \label{fig:ablation}
\vspace{-3mm}
\end{figure}

\begin{figure}
\begin{subfigure}[b]{0.23\textwidth}
        \centering
        \includegraphics[height=1.in]{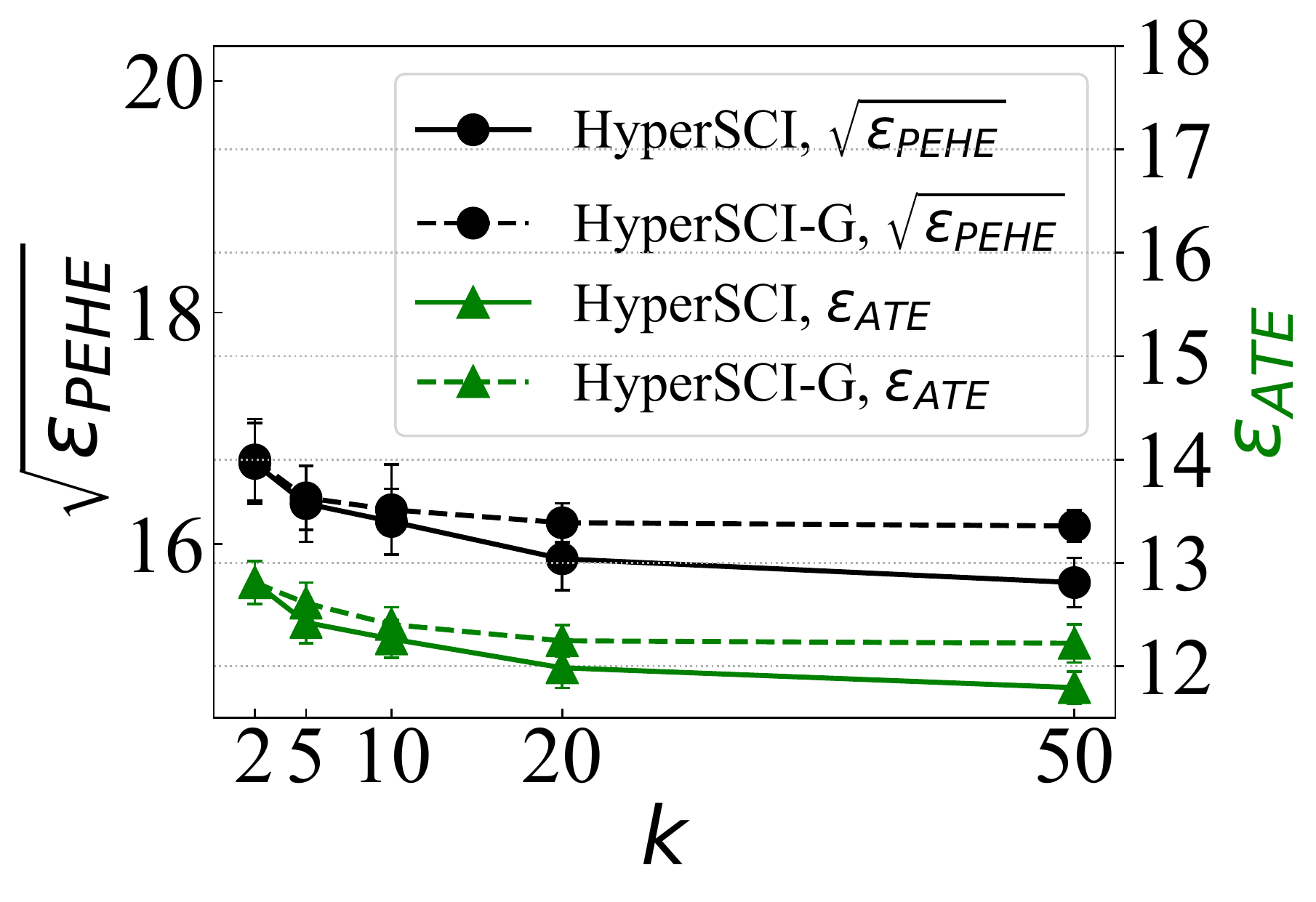}
        \caption{GoodReads, Linear}
    \end{subfigure}
  \begin{subfigure}[b]{0.23\textwidth}
        \centering
        \includegraphics[height=1.in]{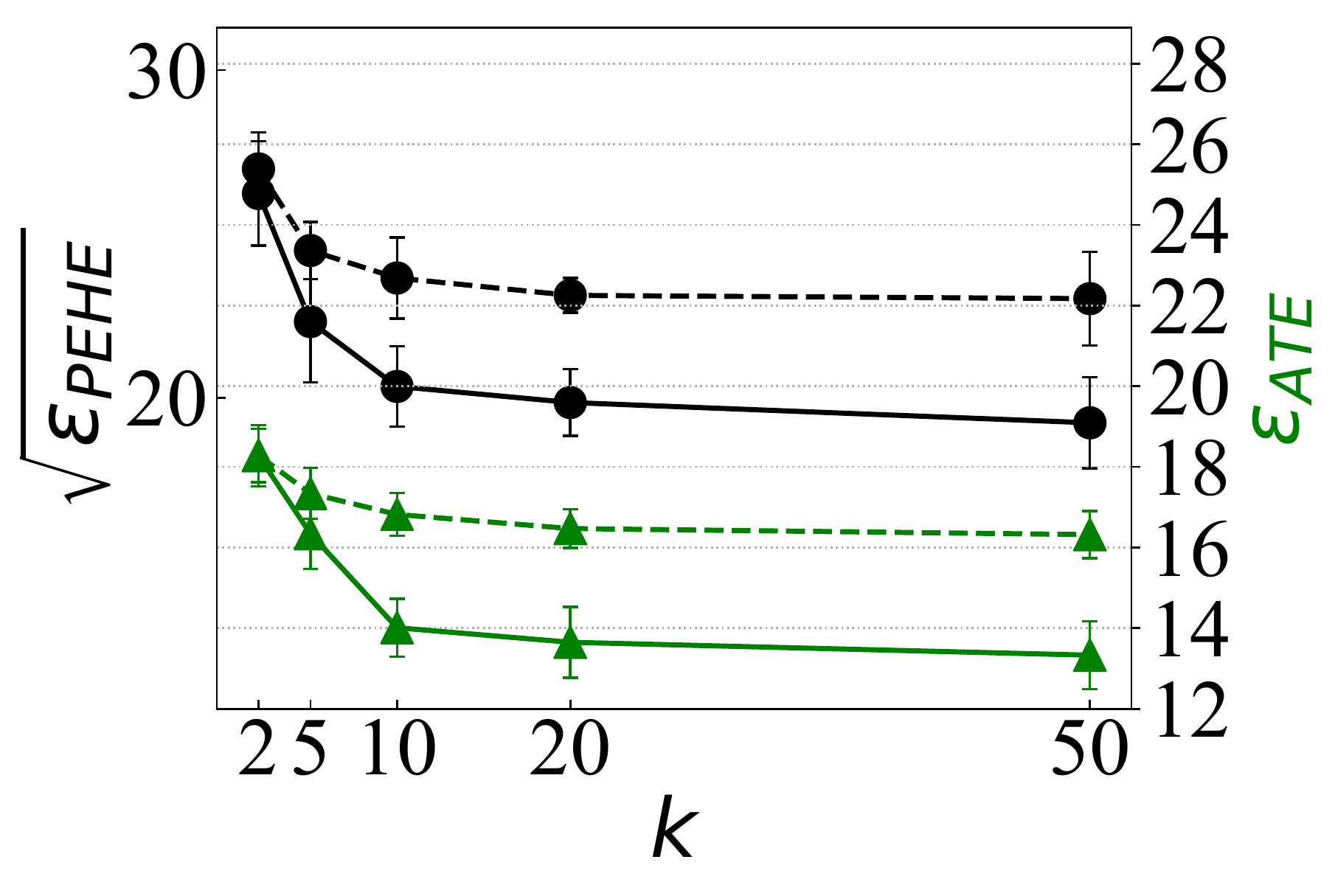}
        \caption{GoodReads, Quadratic}
    \end{subfigure}
    \vspace{-2mm}
  \caption{ITE estimation performance of \mymodel~/ \mymodel-G on hypergraphs with hyperedge size no more than $k$.}
  \label{fig:hyper_k}
\vspace{-1mm}
\end{figure}

\begin{figure}
\begin{subfigure}[b]{0.235\textwidth}
        \centering
        \includegraphics[height=0.9in]{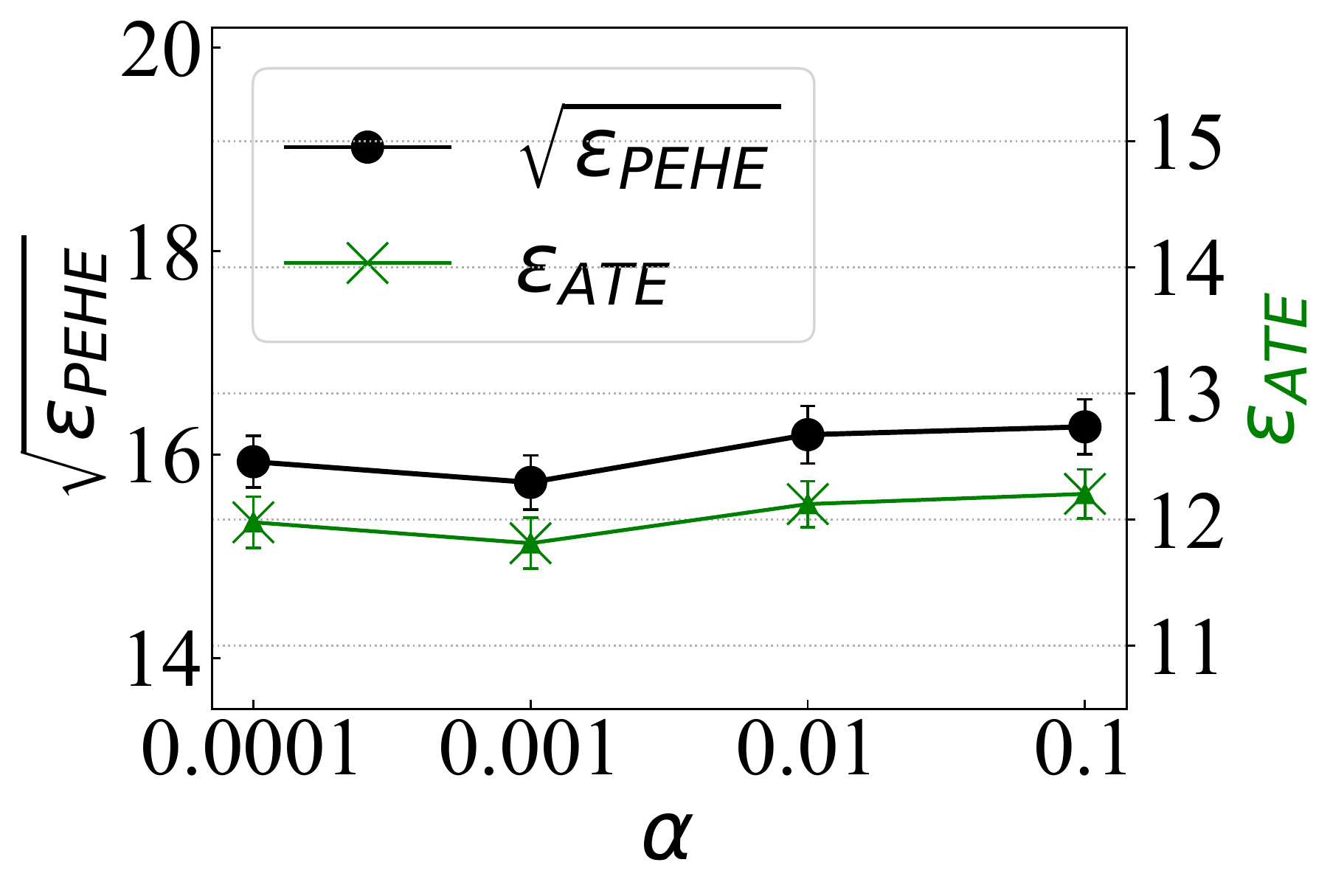}
        \caption{Balancing weight}
    \end{subfigure}
  \begin{subfigure}[b]{0.235\textwidth}
        \centering
        \includegraphics[height=.9in]{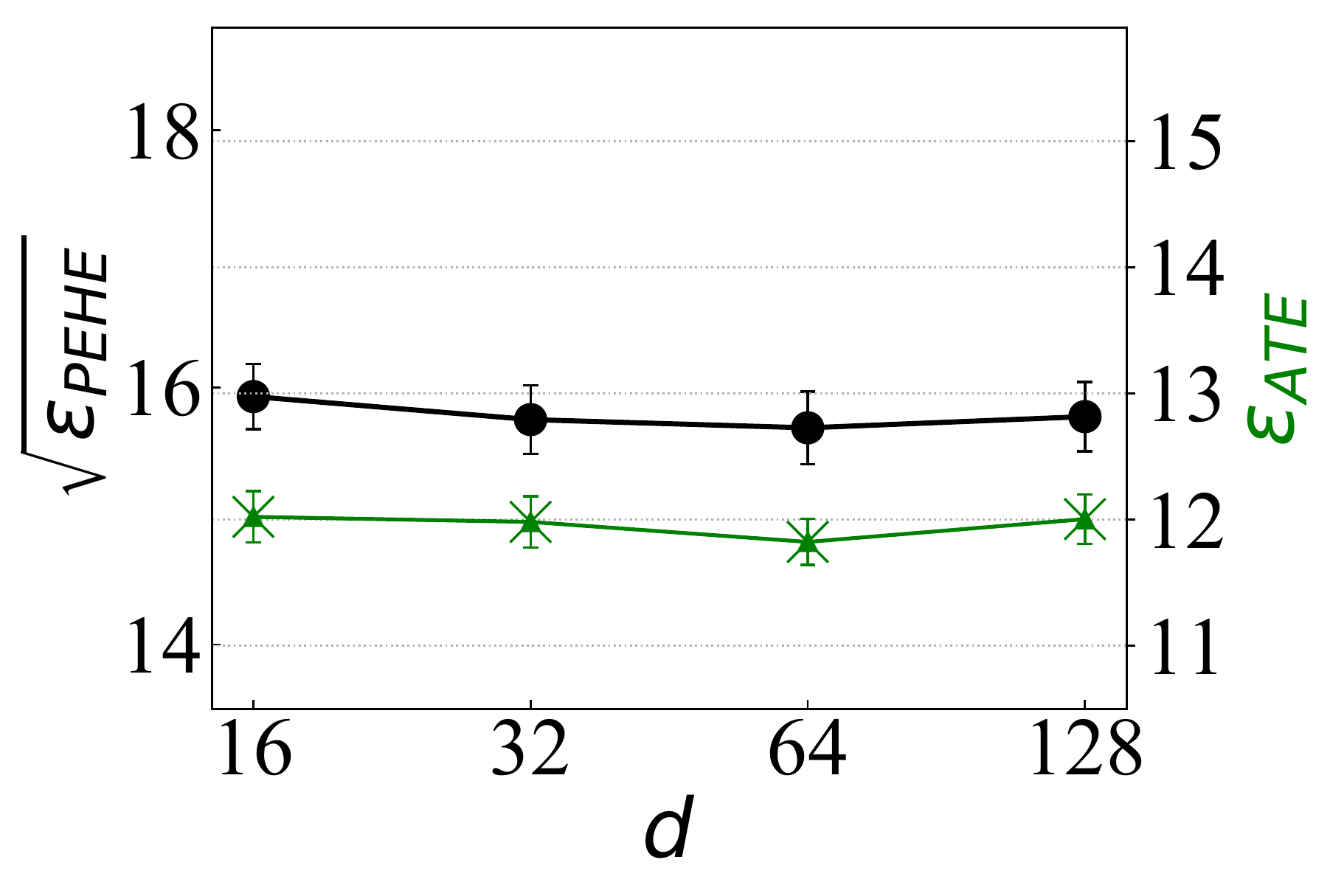}
        \caption{Representation dimension}
    \end{subfigure}
    \begin{subfigure}[b]{0.235\textwidth}
        \centering
        \includegraphics[height=.9in]{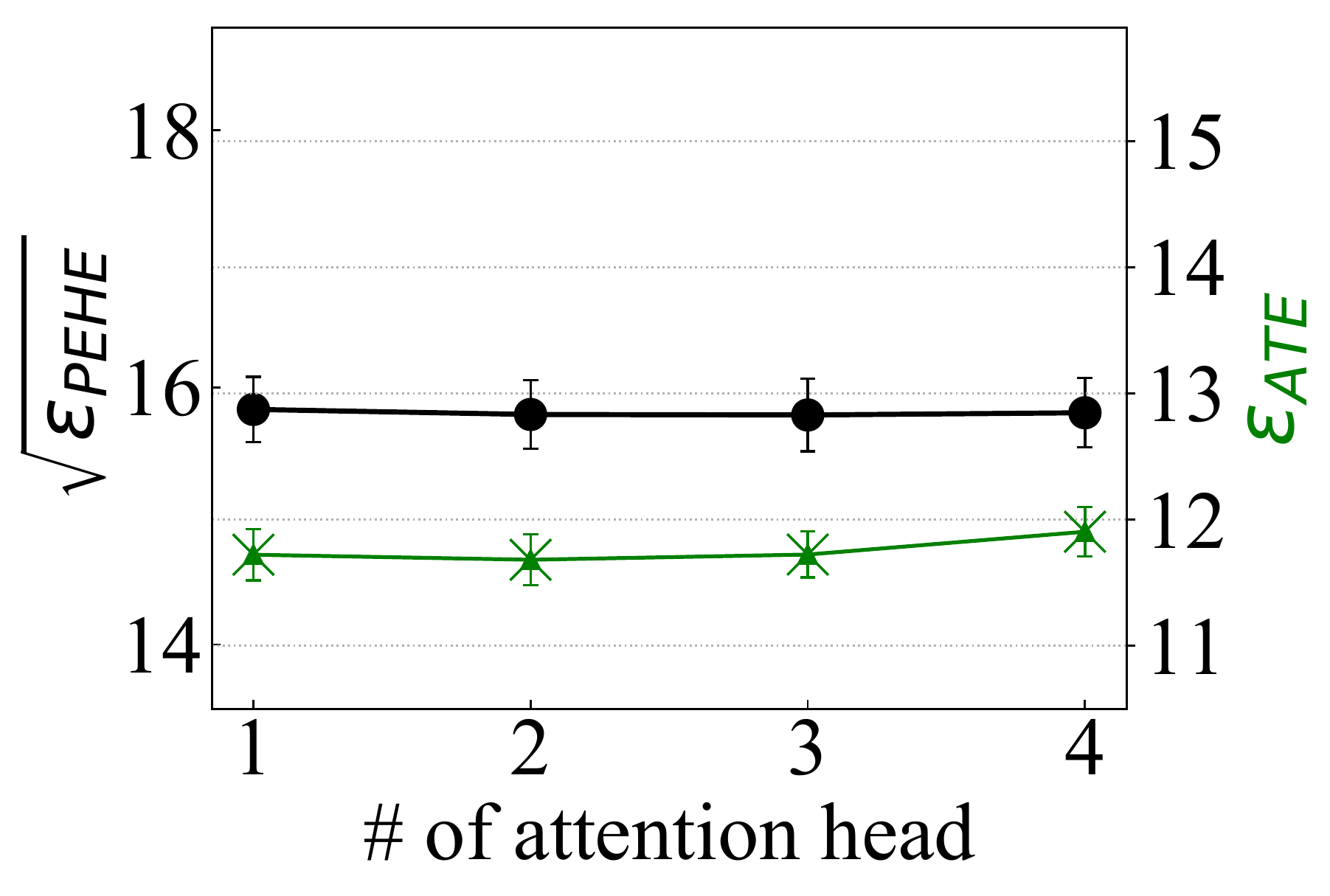}
        \caption{\# of attention head}
    \end{subfigure}
  \begin{subfigure}[b]{0.235\textwidth}
        \centering
        \includegraphics[height=.9in]{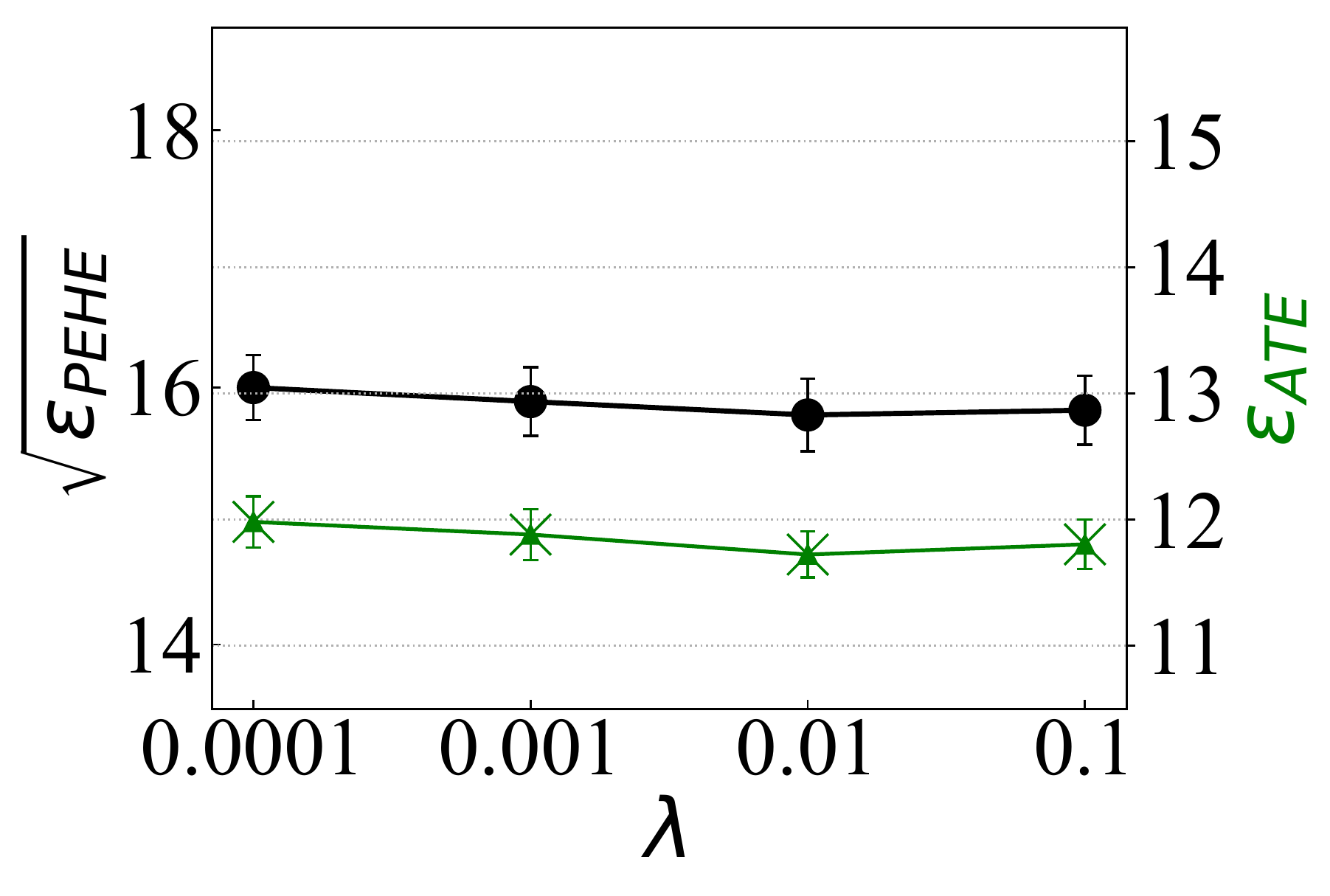}
        \caption{Regularization weight}
    \end{subfigure}
  \caption{ITE estimation performance (mean and standard error) of the proposed framework \mymodel~ under different parameters or model structures on GoodReads dataset.}
  \label{fig:param}
\end{figure}

To investigate the effectiveness of different components in the proposed framework, we conduct ablation studies by considering the following variants: 1) we apply the proposed model \mymodel~ on the projected graph (in a hypergraph structure) (denoted as \textbf{\mymodel-P}); 2) we replace the hypergraph neural network module with a graph neural network module with the same number of layers, and then apply it on the projected graph (in an original graph structure) (\textbf{\mymodel-G}). { Notice that although both evaluated on the projected graph, \textbf{\mymodel-G} handles ordinary graphs with its graph neural network module, while \textbf{\mymodel-P} handles hypergraphs with its hypergraph neural network module}; 3) we remove the balancing techniques in the framework (\textbf{\mymodel-NB}). The ITE estimation results are reported in Fig.~\ref{fig:ablation}, where we notice significant performance gaps between \textbf{\mymodel-P}/\textbf{\mymodel-G} and \mymodel, which imply the effectiveness of modeling the high-order relationships on hypergraphs. We also observe the ITE estimation performance degrades after removing the representation balancing modules, which indicates the effectiveness of the representation balancing techniques on mitigating the biases of ITE estimations.

\vspace{-2mm}
\subsection{A Closer Look at High-Order Interference}
In addition to the overall ITE estimation performance, we take a closer look at high-order interference. 
We investigate how the proposed framework responds to hyperedges with difference sizes. 
More specifically, we remove the hyperedges with size larger than $k$, denote the modified hypergraph as $\mathcal{H}^{(k)}$, and vary the value of $k$. In Fig~\ref{fig:hyper_k}, we compare the ITE estimation performance of the proposed framework \mymodel~ with its variant on the projected ordinary graph \textbf{\mymodel-G}. We observe that: { 1) When $k=2$ (hyperedge size $\le$ 2), the performance of HyperSCI-G is close to HyperSCI. Because when $k=2$, graph convolution can be regarded as a special case of hypergraph convolution with small differences in the graph Laplacian matrix (as illustrated in \cite{bai2021hypergraph}). Empirically this leads to a minor performance difference between HyperSCI-G and HyperSCI;} 2) When $k$ increases, the performance of ITE estimation from both methods are gradually improved, but such an improvement becomes less significant when $k$ is larger. Besides, we notice \mymodel~ consistently outperforms \textbf{\mymodel-G} and such a difference becomes larger as $k$ increases, indicating its efficacy on modeling high-order interference especially on large hyperedges.

\vspace{-1mm}
\subsection{Sensitivity Analysis}

To evaluate the robustness of the proposed framework, we present the ITE estimation performance of \mymodel~under different settings of model hyper-parameters in Fig.~\ref{fig:param}. More specifically, we vary the value of the balancing weight from $\{0.0001,0.001,0.01,0.1\}$, and vary the representation dimension from $\{16, 32, 64, 128\}$. We also vary the number of attention head from $\{1,2,3,4\}$, then change the parameter of regularization weight from $\{0.0001, 0.001, 0.01, 0.1\}$.
As can be observed, our framework is generally robust to different hyper-parameter settings, but proper fine-tuning of these hyper-parameters is still beneficial for the ITE estimation performance. 

%% file: related.tex
\vspace{-2mm}
\section{Related Work}
\vspace{-1mm}
\xhdr{Causal studies under network interference.} There have been many causal studies \cite{aronow2017estimating,basse2018analyzing,imai2020causal,kohavi2013online,tchetgen2012causal,ugander2013graph,yuan2021causal,ma2021causal,bhattacharya2020causal} which address the existence of network interference. These works mainly include the following categories: (i) \textit{Random assignment strategy under interference} \cite{aronow2017estimating,basse2018analyzing,imai2020causal,ugander2013graph,fatemi2020minimizing}. These works focus on experimental studies under interference (without SUTVA assumption). 
In some studies \cite{karrer2021network}, strong interference is assumed to exist within each group while there is no interference across different groups; (ii) \textit{Causal effect estimation on observational data with interference} \cite{rakesh2018linked,ma2021causal,arbour2016inferring,tchetgen2021auto}. Different from the experimental studies which can design assignment strategy, another line of works (and also our work) assume interferences exist across individuals in the observational data. They relax the SUTVA assumption and define the potential outcome with a function that takes the instance covariates and treatment assignment of each individual and other interacted individuals as input. Among them, Rakesh et al. \cite{rakesh2018linked} propose a Linked
Causal Variational Autoencoder (LCVA) framework to estimate the causal effect of a treatment on an outcome with the existence of interference between pairs of instances. Different from these works that focus on pairwise spillover effects, Ma et al. \cite{ma2021causal} consider the spillover effect in network structure, and propose a graph neural network (GNN) \cite{kipf2016semi} based framework for causal effect estimation under network interference. 
However, these works are still limited in pairs of individuals or ordinary graphs and lack consideration of high-order interference.
{
Another line of studies is bipartite causal inference  \cite{zigler2021bipartite,pouget2019variance}. Traditionally, bipartite causal inference involves two types of units: interventional/outcome units. Interventional units are assigned with treatments, and outcomes are observed from outcome units.  
Although this setup is different from ours, considering that there is a node-hyperedge bipartite corresponding to each hypergraph (and ordinary graph), thus 
the two modeling approaches (bipartite and hypergraph) are conceptually similar. Nevertheless, 
we argue hypergraph is a more appropriate framing in many scenarios since: i) hypergraph does not require instantiating edges as additional nodes, or treating these two kinds of nodes differently, thus more computationally efficient; ii) hypergraph has the potential to be more convenient and efficient when generalizing to new hyperedges, while bipartite needs to generate both new nodes for the new hyperedges and their associated new edges.
}

\xhdr{Hypergraph algorithms and neural networks.} To process hypergraph structures for downstream tasks, a line of works simplify the hypergraph structure by taking abstract representations of complicated multi-way interactions \cite{benson2018simplicial,li2013link,xu2013hyperlink,yadati2018link,zhang2018beyond}. Other works directly tackle the original hypergraph structure \cite{arya2018exploiting,benson2018sequences,feng2019hypergraph,huang2015scalable,sharma2014predicting,yadati2018hypergcn}.  
Recently, numerous works have studied on hypergraph neural networks \cite{bai2021hypergraph,yadati2018hypergcn}. Feng el al. \cite{feng2019hypergraph} propose hypergraph neural networks (HGNN) framework to encode high-order data correlation in a hypergraph structure. A hyperedge convolution operation is designed for representation learning. Bai et al. \cite{bai2021hypergraph} introduce two end-to-end trainable operators hypergraph convolution and hypergraph attention to learn node representations in hypergraphs. Yadati et al. \cite{yadati2018hypergcn} develop a self-attention based hypergraph neural network Hyper-SAGNN, which is applicable to homogeneous or heterogeneous hypergraphs with variable hyperedge sizes. Jiang et al. \cite{jiang2019dynamic} propose a dynamic hypergraph neural network (DHGNN) which can dynamically update hypergraph structure on each layer. 

%% file: appendix.tex
\appendix
\section{Notation Table}
We use Table~ \ref{tab:notation} to list the most important notations used in this paper.
\begin{table}[H]
  \caption{Notation.}
  \label{tab:notation}
  \vspace{-4mm}
  \begin{tabular}{ll}
    \toprule
    Notation & Definition \\ 
    \midrule
    $\mathcal{H}$          & hypergraph\\
    $\mathcal{V},\mathcal{E}$          & the set of nodes/hyperedges \\
    $\mathbf{H}$          & hypergraph structure matrix\\
    $n$               & the number of nodes \\
    $m$               & the number of hyperedges\\
    $\mathcal{N}_i$ & the set of neighboring nodes for the $i$-th node \\
    $\mathcal{N}_e$ & the set of nodes on the hyperedge $e$ \\
    $\mathcal{E}_i$ & the set of hyperedges which contains the $i$-th node\\
    $\mathbf{X},\mathbf{x}_i$ & features of all nodes/the $i$-th node\\
    $\mathbf{T},{t}_i$ & treatment assignment of all nodes/the $i$-th node\\
    $\mathbf{Y},{y}_i$ & observed outcome of all nodes/the $i$-th node\\
    ${y}^1_i,y_i^0$ & potential outcomes of the $i$-th node\\
    $\Phi_Y(\cdot)$ & potential outcome function \\
    $\tau_i,\hat{\tau}_i$ & true/predicted ITE for the $i$-th node\\
    $y_i, \hat{y}_i$ & true/predicted outcome for the $i$-th node\\
    $(\cdot)_{-i}$ & variables for all the nodes except the $i$-th node\\
    $\delta_i$ & the spillover effect of the $i$-th node\\
    $\mathtt{SMR}(\cdot)$ & summary function\\
    $\mathbf{O}, \mathbf{o}_i$ & the environment information of the $i$-th node\\
    $\mathbf{Z},\mathbf{z}_i$ & confounder representations of all nodes/the $i$-th node\\
    $\Psi(\cdot)$ & interference representation learner\\
    $f_1(\cdot),f_0(\cdot)$ & potential outcome prediction functions\\
    $\mathbf{P},\mathbf{p}_i$ & interference representations of all nodes/the $i$-th node\\
    $r(i)$ & the ratio of the treatment assignment of the $i$-node \\&in its neighborhood\\
    \bottomrule
\end{tabular}
\end{table}

\section{More Experimental Results}
\subsection{ITE Estimation Performance under Different Settings on All the Datasets}
\begin{figure*}
\centering
  \begin{subfigure}[b]{0.235\textwidth}
        \centering
        \includegraphics[height=0.95in]{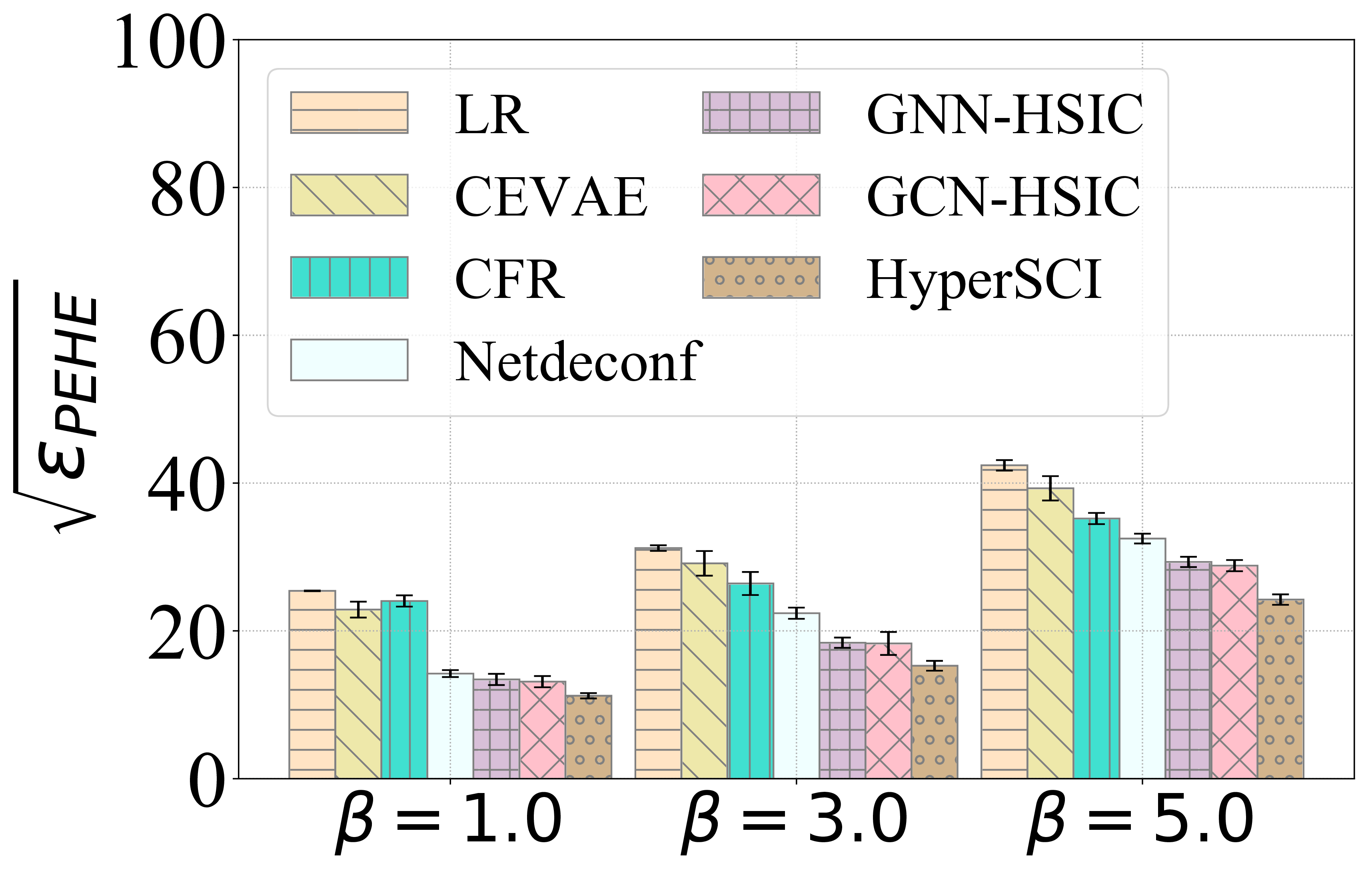}
        \caption{Linear, $\sqrt{\epsilon_{PEHE}}$, CT}
    \end{subfigure}
  \begin{subfigure}[b]{0.235\textwidth}
        \centering
        \includegraphics[height=0.95in]{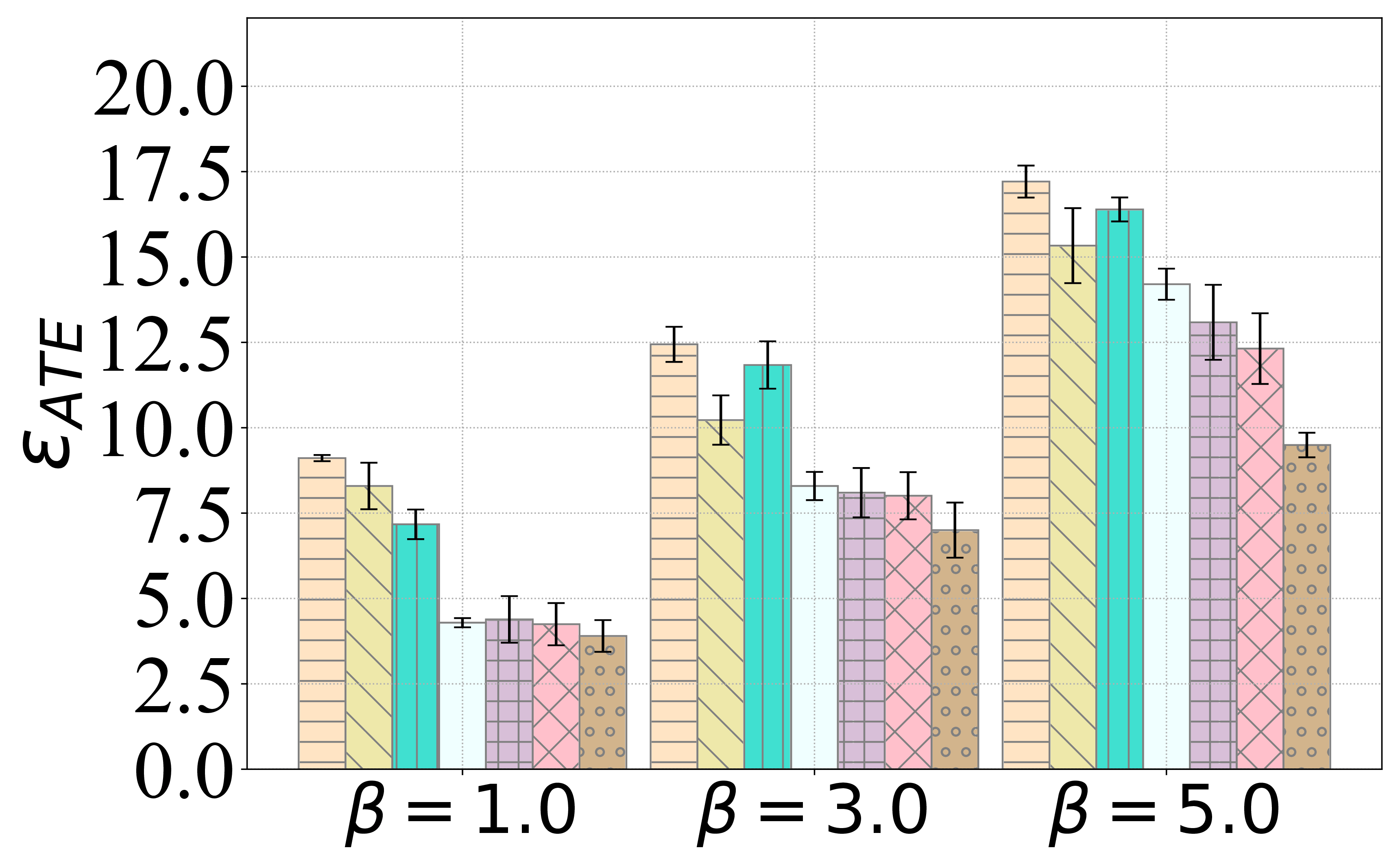}
        \caption{Linear, $\epsilon_{ATE}$, CT}
    \end{subfigure}
  \begin{subfigure}[b]{0.235\textwidth}
        \centering
        \includegraphics[height=0.95in]{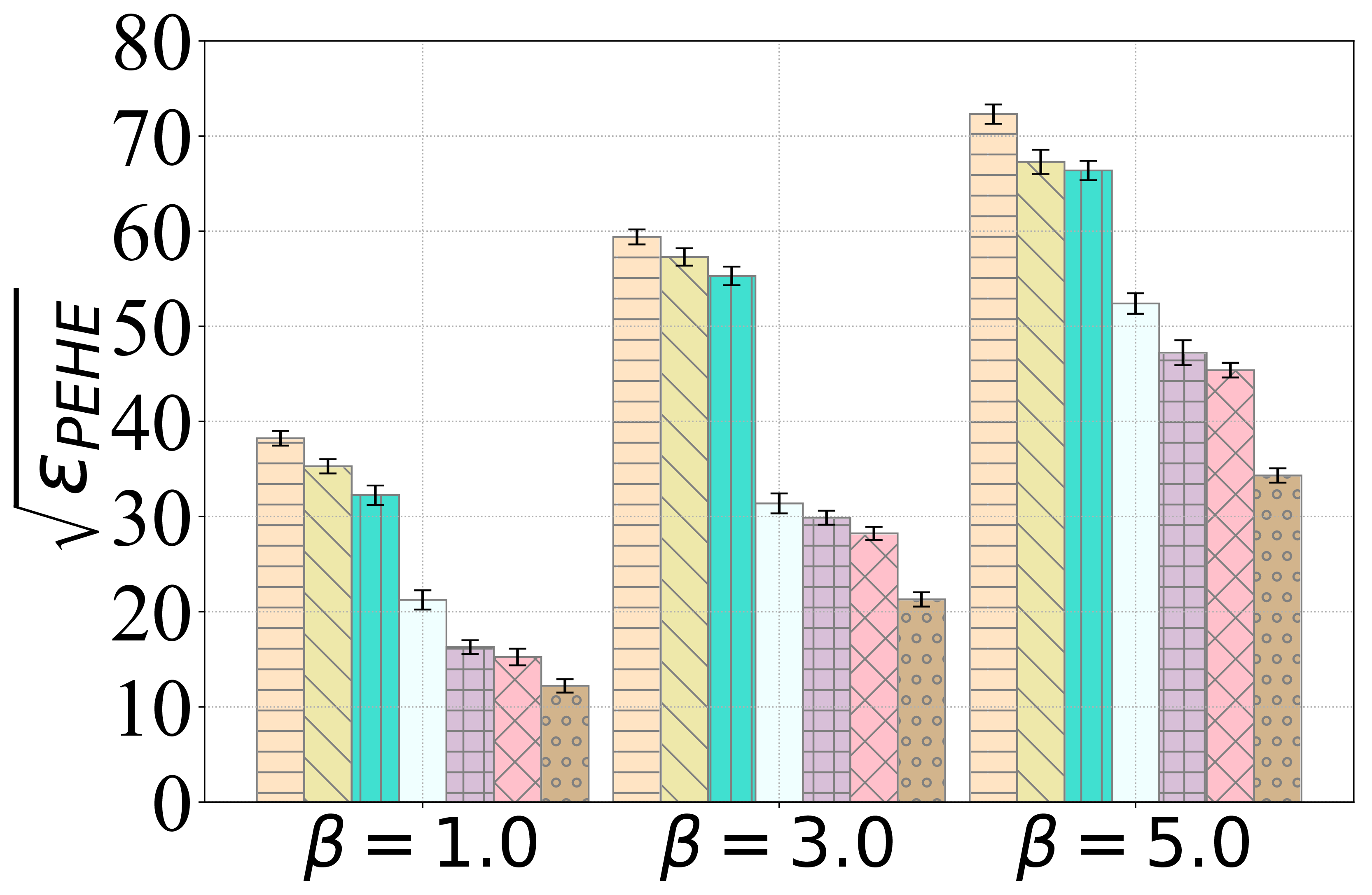}
        \caption{Quadratic, $\sqrt{\epsilon_{PEHE}}$, CT}
    \end{subfigure}
    \begin{subfigure}[b]{0.235\textwidth}
        \centering
        \includegraphics[height=0.95in]{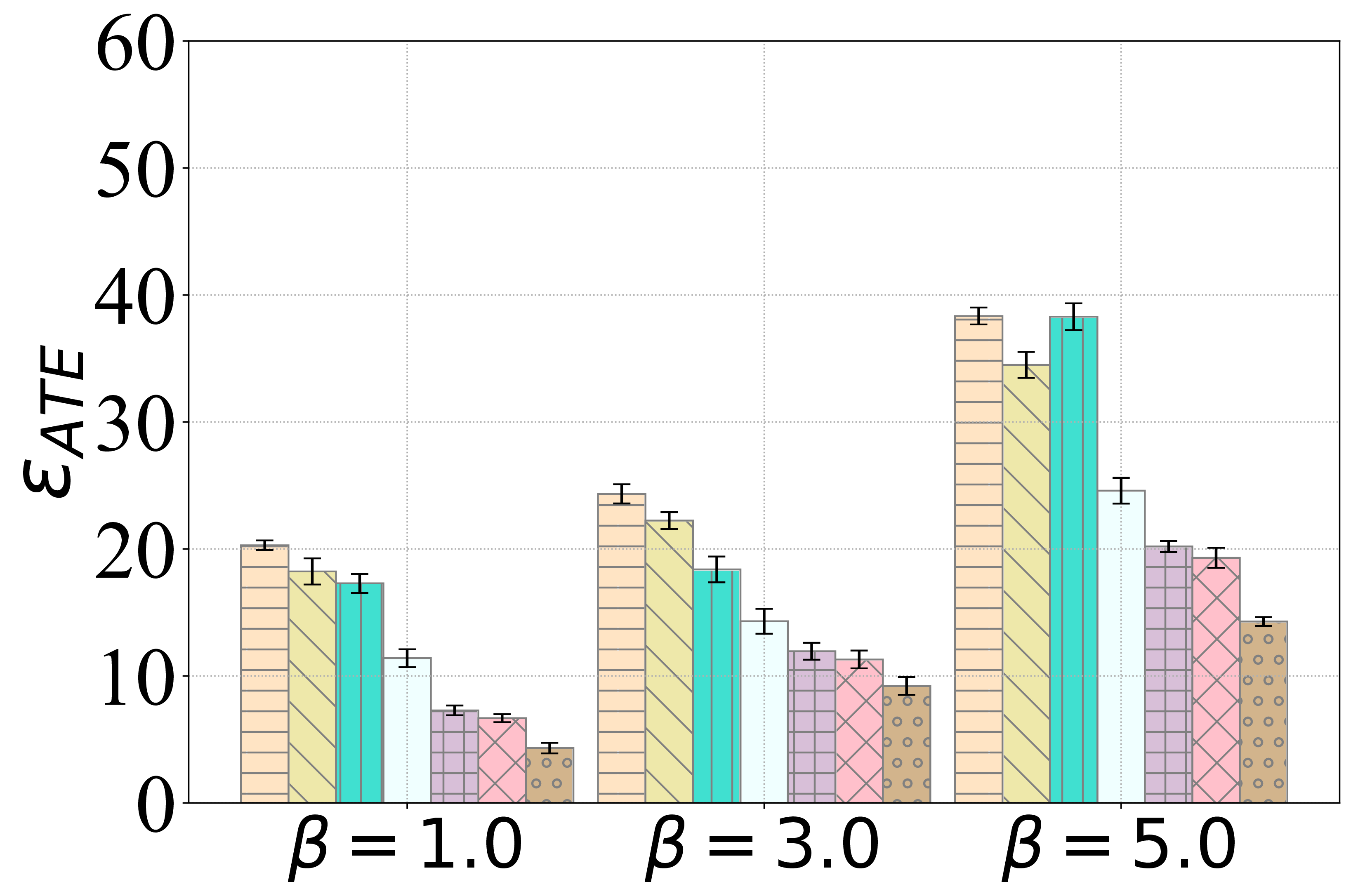}
        \caption{Quadratic, $\epsilon_{ATE}$, CT}
    \end{subfigure}
    \begin{subfigure}[b]{0.235\textwidth}
        \centering
        \includegraphics[height=0.95in]{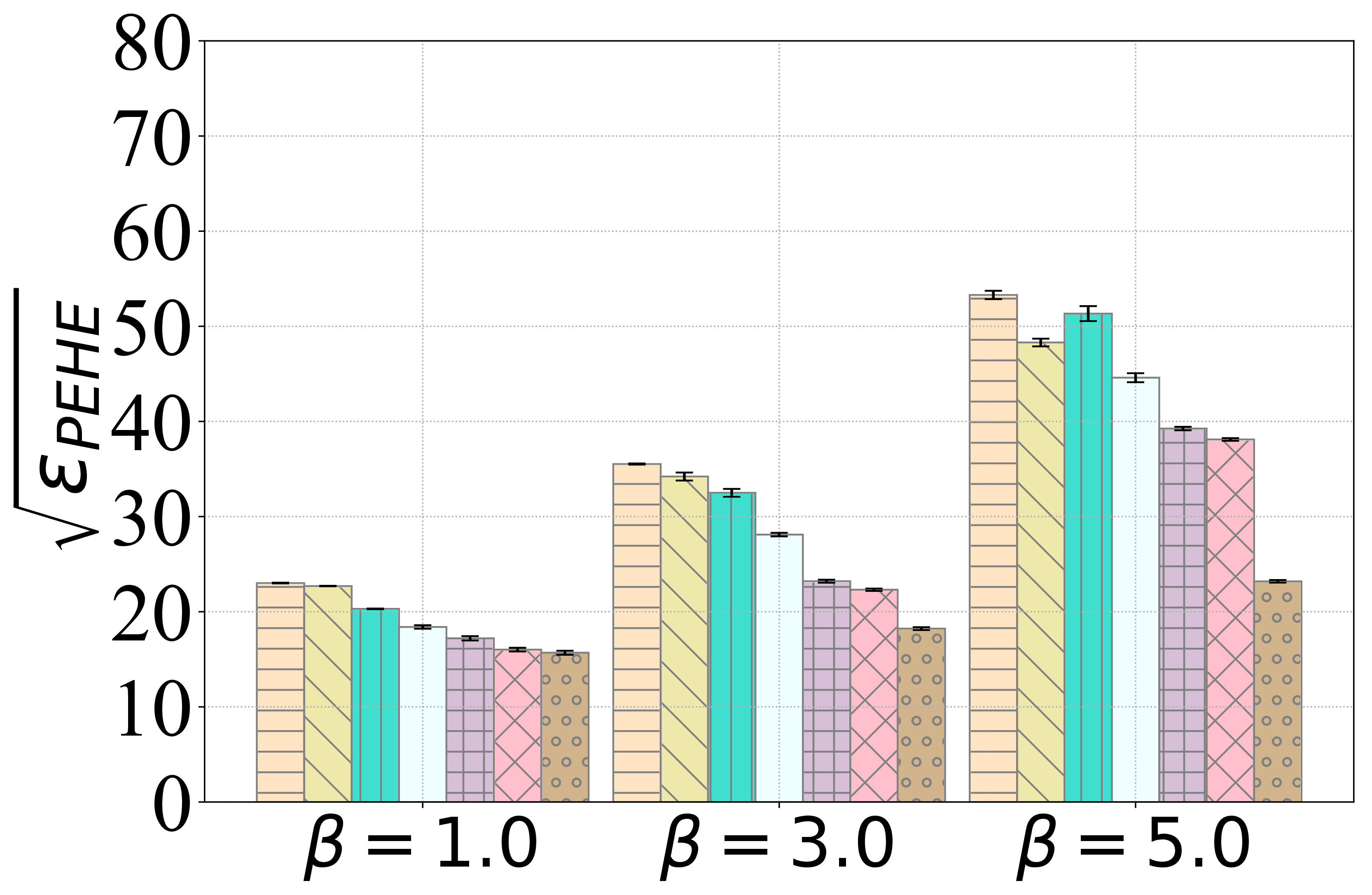}
        \caption{Linear, $\sqrt{\epsilon_{PEHE}}$, GR}
    \end{subfigure}
  \begin{subfigure}[b]{0.235\textwidth}
        \centering
        \includegraphics[height=0.95in]{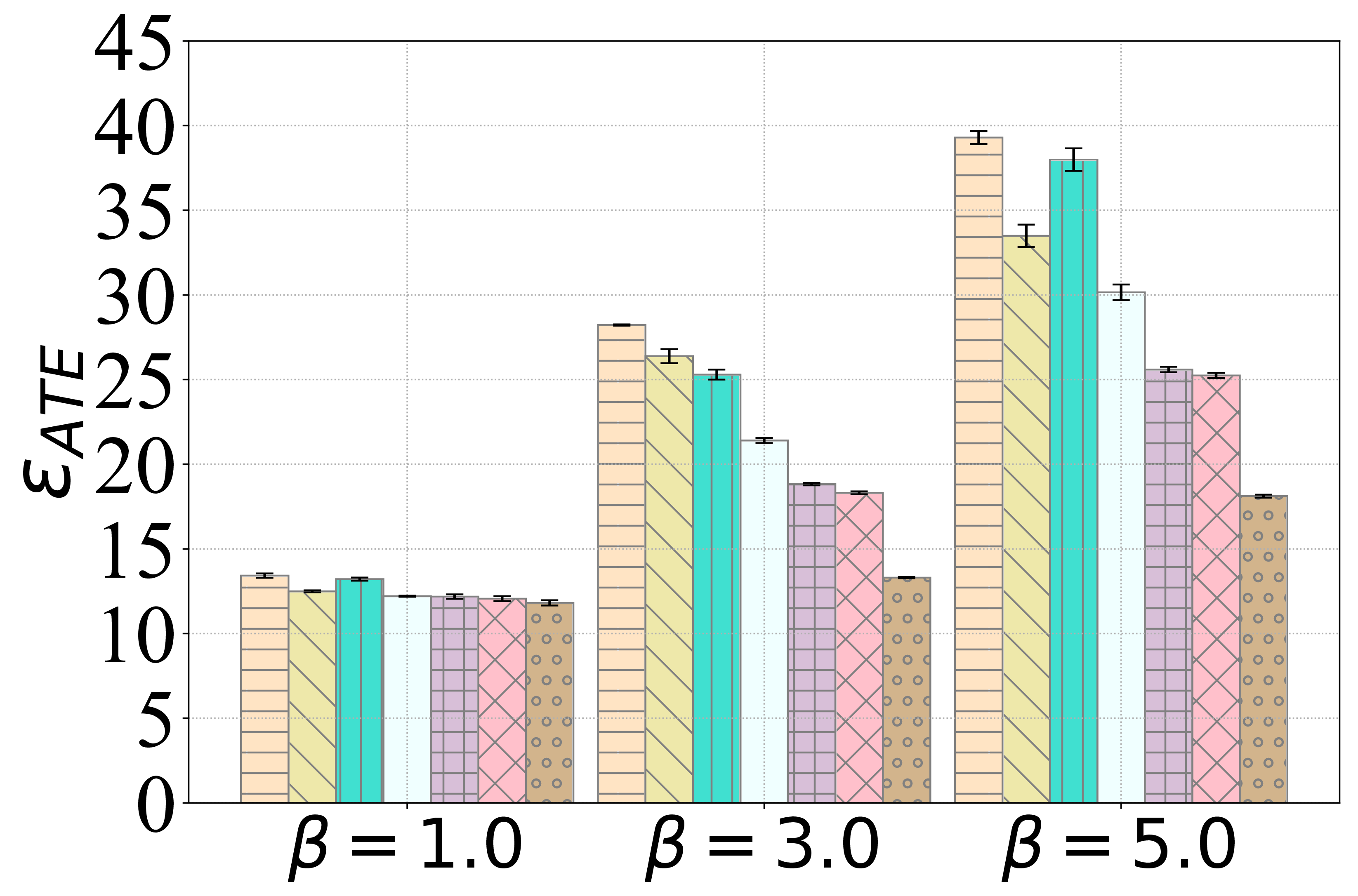}
        \caption{Linear, $\epsilon_{ATE}$, GR}
    \end{subfigure}
  \begin{subfigure}[b]{0.235\textwidth}
        \centering
        \includegraphics[height=0.95in]{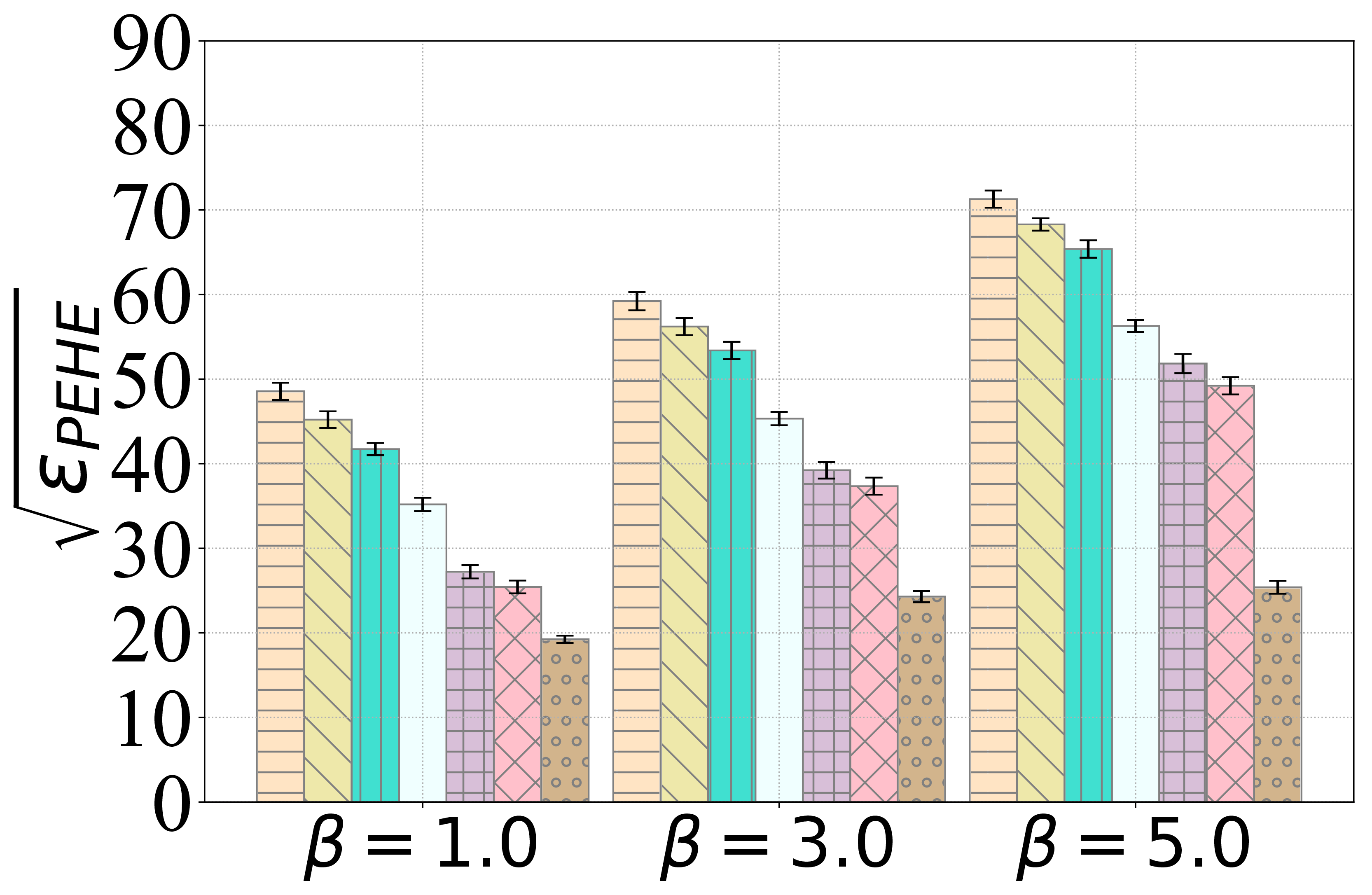}
        \caption{{\small Quadratic,$\sqrt{\epsilon_{PEHE}}$,GR}}
    \end{subfigure}
    \begin{subfigure}[b]{0.235\textwidth}
        \centering
        \includegraphics[height=0.95in]{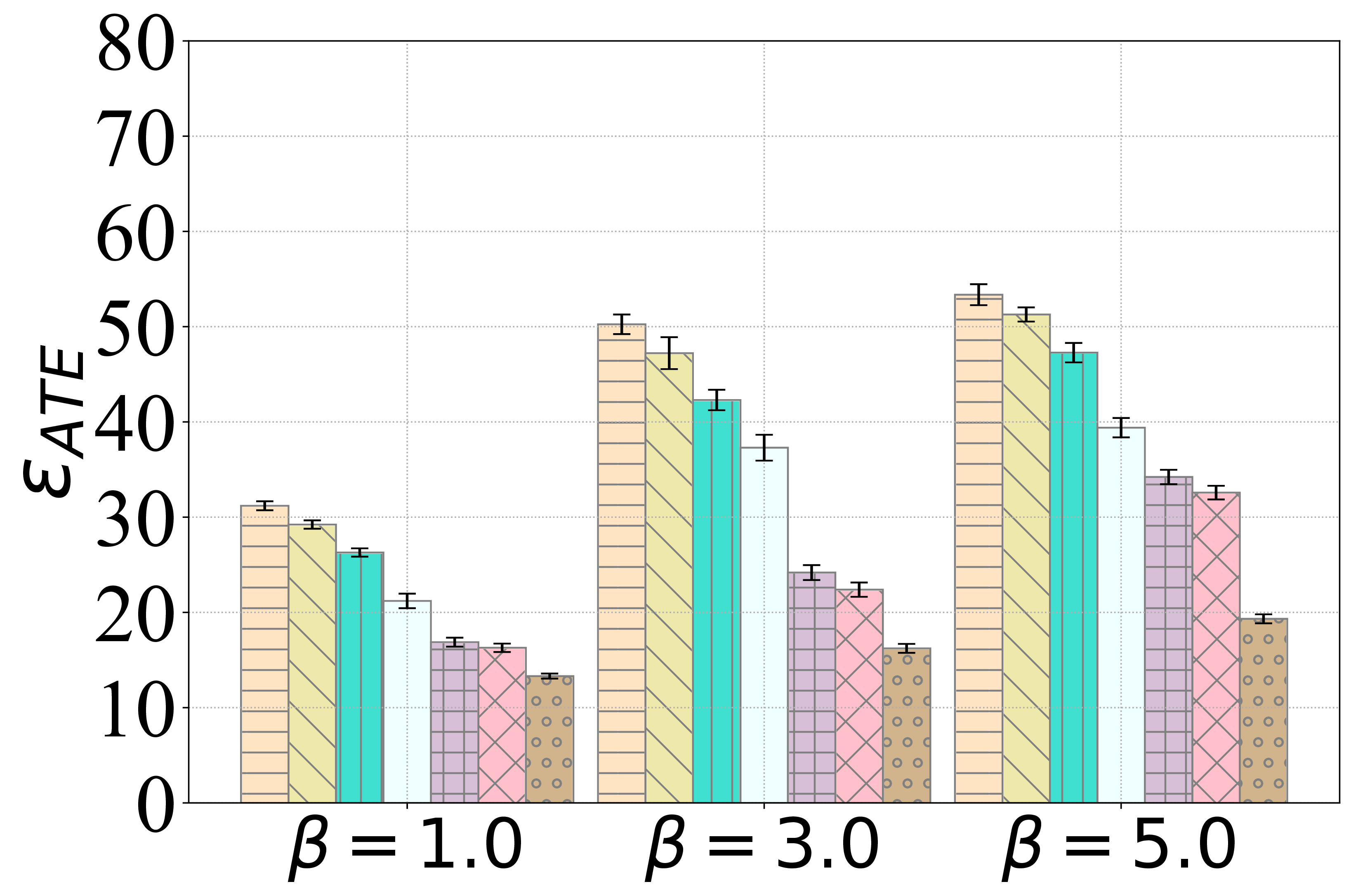}
        \caption{Quadratic, $\epsilon_{ATE}$, GR}
    \end{subfigure}
    \begin{subfigure}[b]{0.235\textwidth}
        \centering
        \includegraphics[height=0.95in]{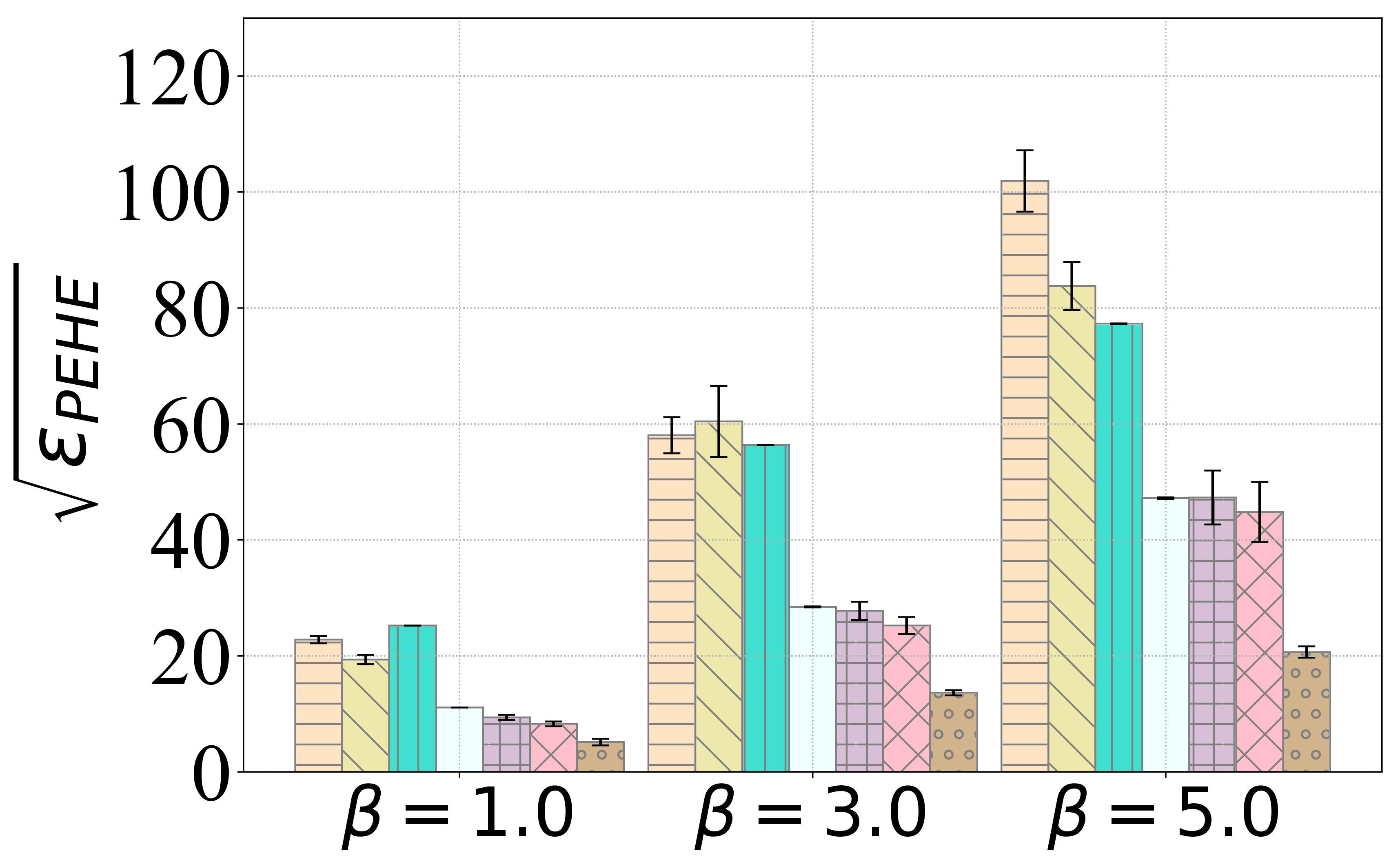}
        \caption{Linear, $\sqrt{\epsilon_{PEHE}}$, MS}
    \end{subfigure}
  \begin{subfigure}[b]{0.235\textwidth}
        \centering
        \includegraphics[height=0.95in]{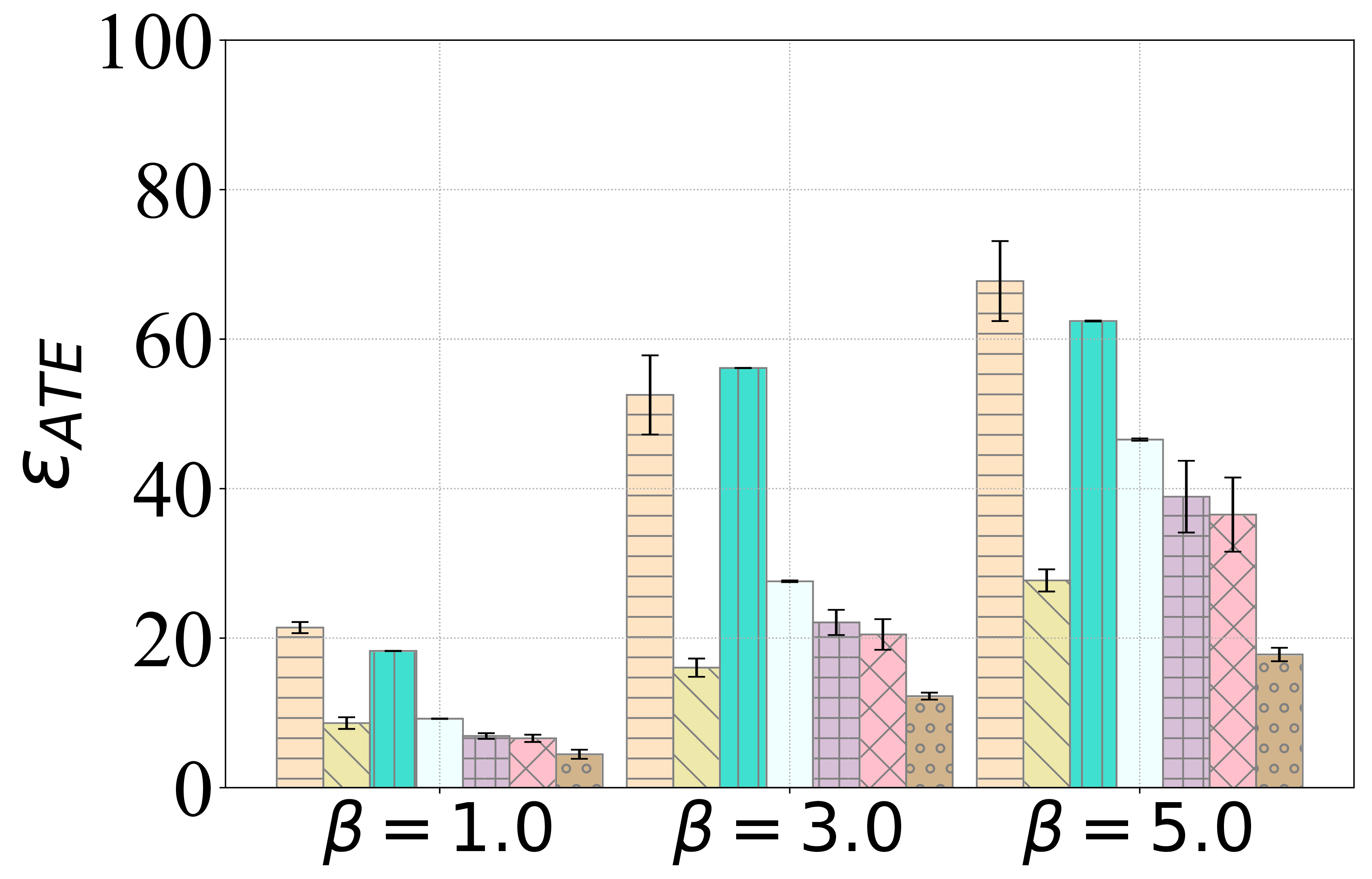}
        \caption{Linear, $\epsilon_{ATE}$, MS}
    \end{subfigure}
  \begin{subfigure}[b]{0.235\textwidth}
        \centering
        \includegraphics[height=0.95in]{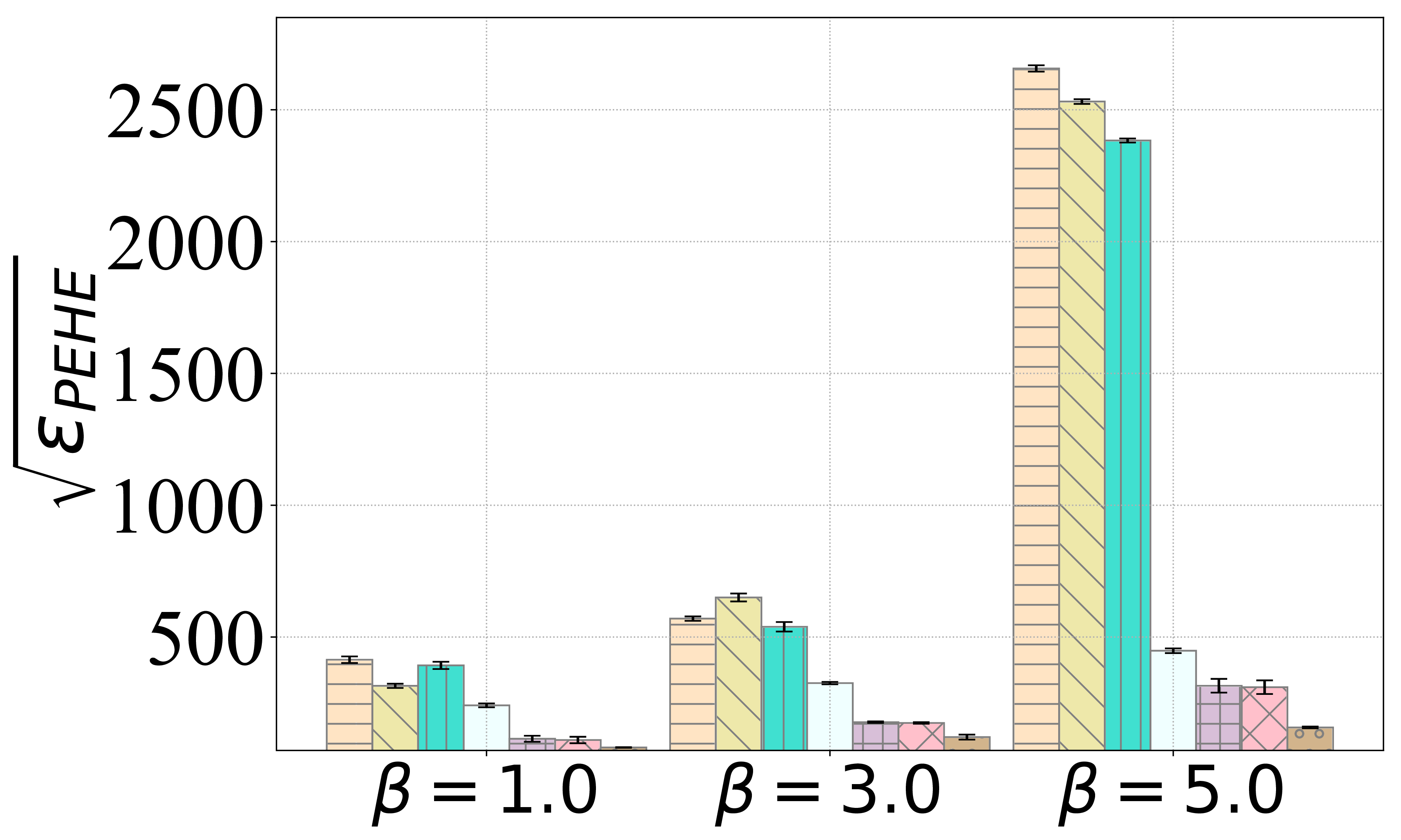}
        \caption{Quadratic, $\sqrt{\epsilon_{PEHE}}$, MS}
    \end{subfigure}
    \begin{subfigure}[b]{0.235\textwidth}
        \centering
        \includegraphics[height=0.95in]{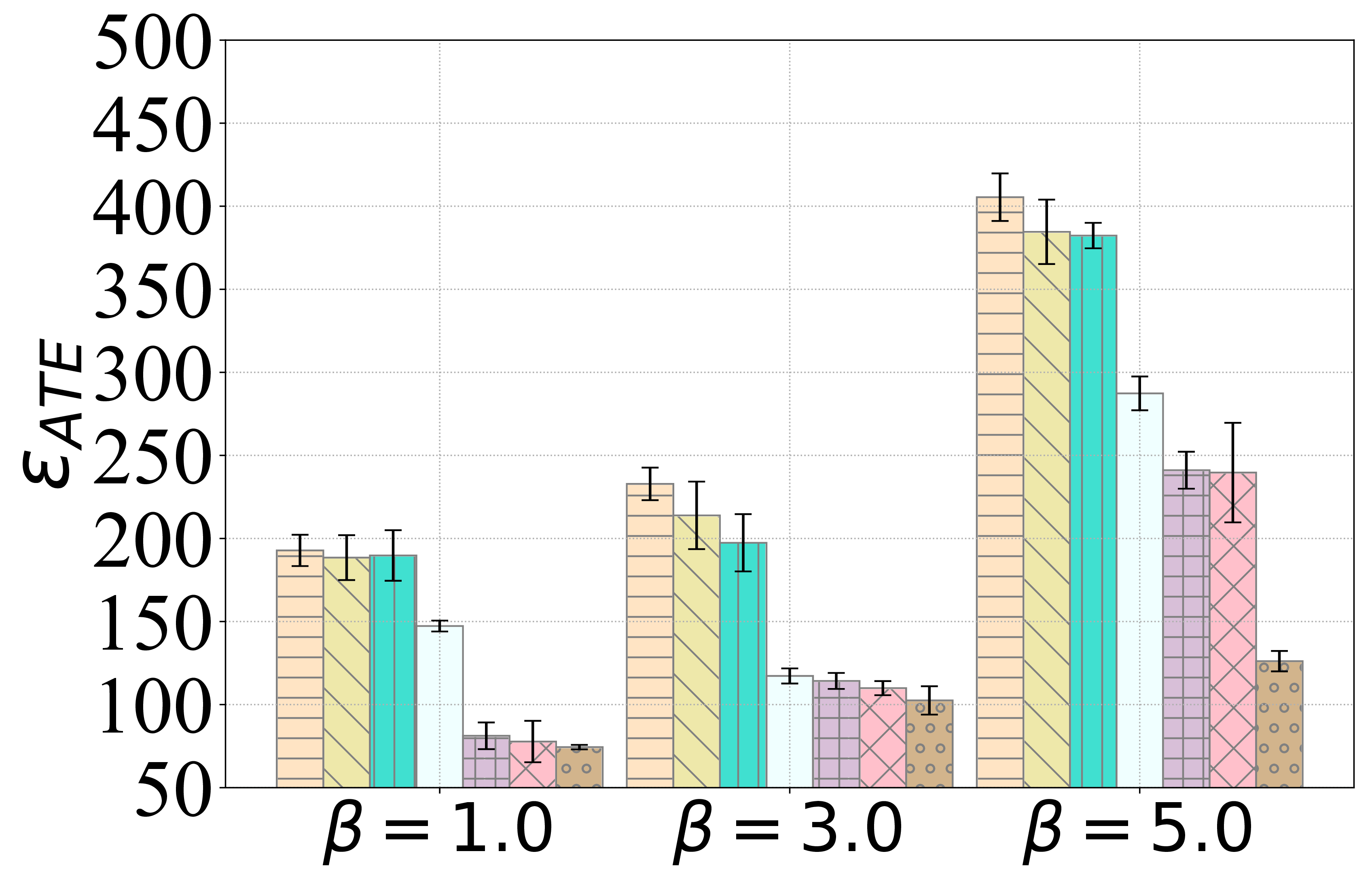}
        \caption{Quadratic, $\epsilon_{ATE}$, MS}
    \end{subfigure}
  \caption{Comparison of the performance of ITE estimation under different settings. ``CT", ``GR" and ``MS" stand for Contact, GoodReads and Microsoft Teams datasets, respectively.}
  \label{fig:ITE_complete}
\vspace{-3mm}
\end{figure*}

In this section, we show the ITE estimation performance under different settings (including the linear and quadratic settings with $\beta$ among $\{1.0, 3.0, 5.0\}$) on all the datasets in Fig.~\ref{fig:ITE_complete}. We can observe that the proposed \mymodel~ consistently outperforms the baselines under different settings on all the datasets. The superiority of our framework against baselines becomes more obvious when $\beta$ is larger (i.e., the interference is stronger), because our framework can better handle the interference in the hypergraph.



\subsection{Case Studies}
\begin{figure}[H]
  \begin{subfigure}[b]{0.207\textwidth}
        \centering
        \includegraphics[height=1.1in]{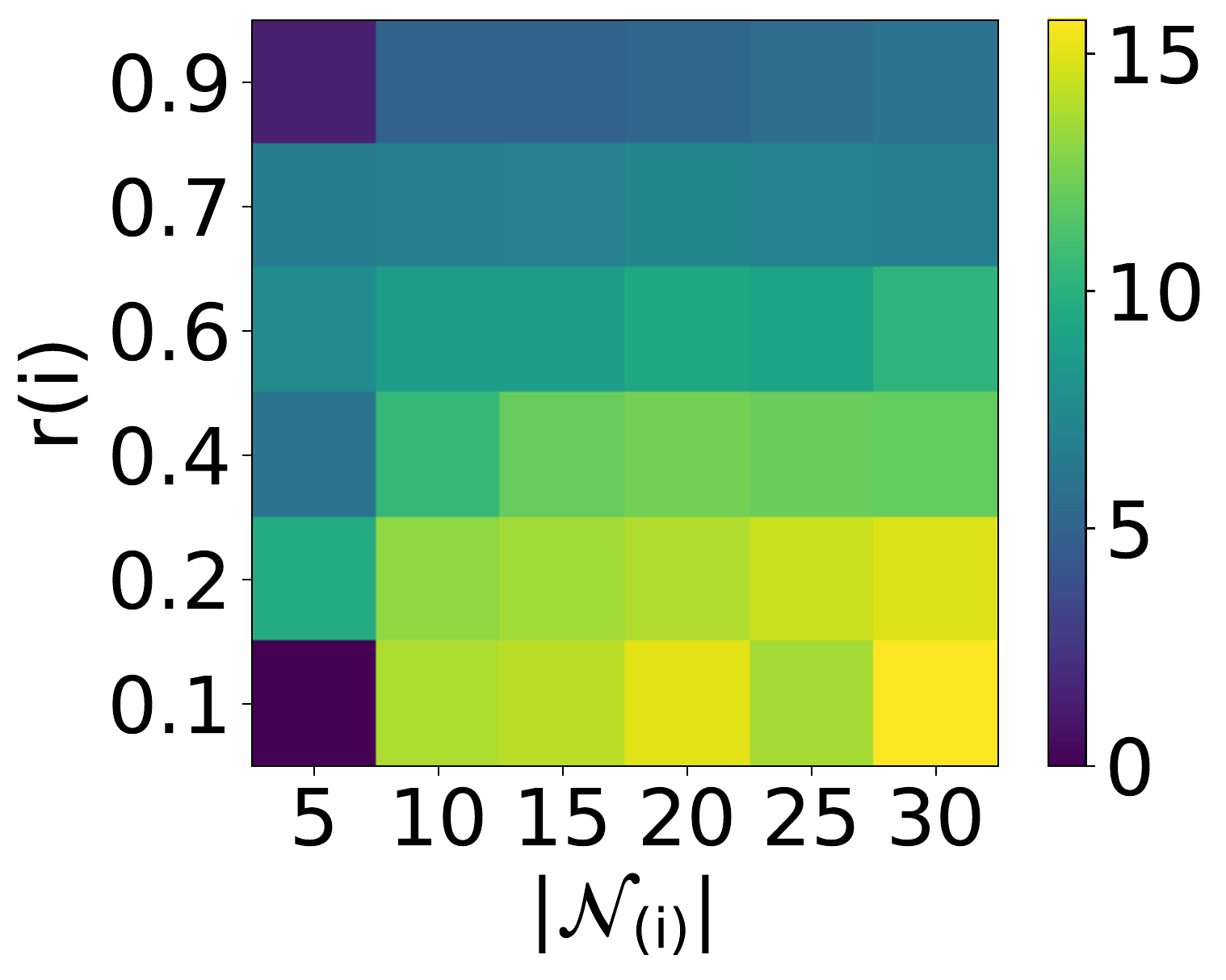}
        \caption{Heatmap}\label{fig:heatmap}
    \end{subfigure}
    ~
    \begin{subfigure}[b]{0.29\textwidth}
        \centering
        \includegraphics[height=1.1in]{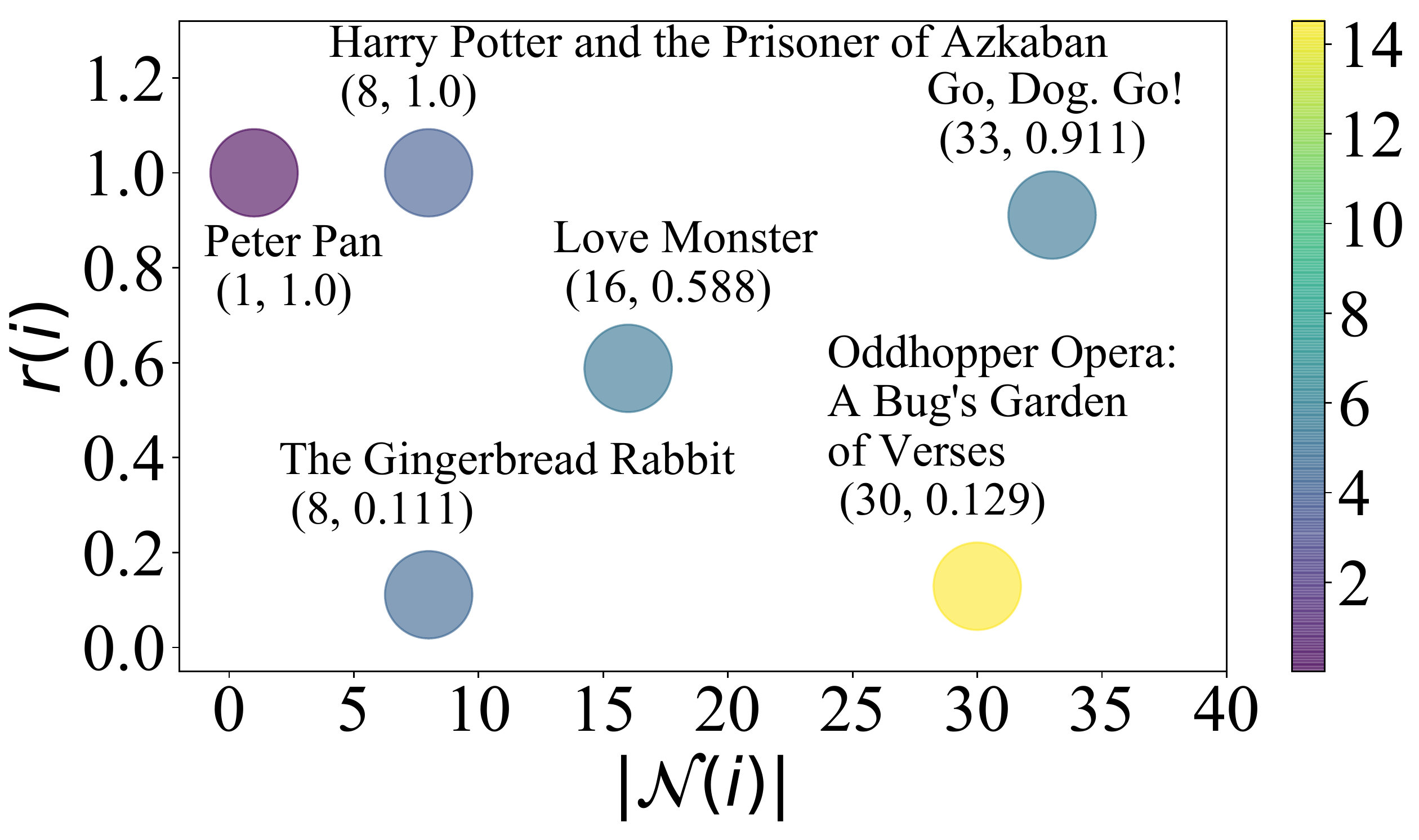}
        \caption{Case studies}\label{fig:case-study}
    \end{subfigure}
    \vspace{-4mm}
  \caption{
  (a) Heatmap: the difference between ITE estimations with hypergraph and with  projected ordinary graph on GoodReads. Nodes are divided into $6\times 6$ grids w.r.t. their number of neighbors $|\mathcal{N}_i|$ and the homophily of treatment assignment $r(i)$. 
  (b) Case studies of representative books. 
  }
\end{figure}

We further conduct case studies to investigate how the proposed method acts on individuals in responding to their neighboring nodes (i.e., the size of one's neighborhood and the homophily of treatment assignments within one's neighborhood). The neighborhood of $i$ is defined as the set of nodes which are connected with $i$ via any hyperedges, i.e., $\mathcal{N}_i=\bigcup_{e\in\mathcal{E}_i}\{j\in\mathcal{N}_e\}$. The homophily of treatment assignment is defined as the ratio of neighboring nodes which share the same treatment assignment as oneself, i.e., $r(i)=\frac{\sum_{j\in \mathcal{N}_i}\mathbf{1}(t_j=t_i)}{|\mathcal{N}_i|}$. In Fig.~\ref{fig:heatmap}, we show the difference between the ITE estimation results made 
with the original hypergraph and with the projected graph, w.r.t. $\mathcal{N}_i$ and $r(i)$. Overall, we see larger divergences on individuals with a larger neighborhood size but less agreement with their neighbors in terms of treatment assignments. 
In Fig.~\ref{fig:case-study}, we further showcase the insights by presenting several representative children books on the GoodReads dataset. For example, the author of ``Peter Pan'' had not published many works but these books all received good rating scores, leading to a ``consistent" reputation of the author. Therefore, the outcome of the book ``Peter Pan'' is less impacted by the high-order interference among its neighbors. 
On the other hand, the high rating score of the book ``Oddhopper Opera'' differs from most of its neighbors, leading to a mixed reputation of the author. In this case, the potential outcome is more likely to be affected by the high-order interference on the hypergraph.

\section{Details of Experimental Settings}
All the experiments are conducted under the following environment:
\begin{itemize}
    \item Operating system: Ubuntu 18.04
    \item GPU memory: 16GB
    \item Pytorch 1.9.0, Cuda ToolKit 11.1, cuDNN 8.0.5
\end{itemize}

\noindent\textbf{Baseline parameter settings.} For the baselines CEVAE, CFR, Netdeconf, GNN-HSIC, and GCN-HSIC, we set the representation dimension as 32, 32, 100, 64, respectively. The numbers of training epochs for these baselines are set as $500$. The number of samples in CEVAE in training is set as $5$ by default.

